\documentclass[lettersize,journal]{IEEEtran}
\usepackage{amsmath,amsfonts}
\usepackage{algorithmic}
\usepackage{algorithm}
\usepackage{array}
\usepackage[caption=false,font=normalsize,labelfont=sf,textfont=sf]{subfig}
\usepackage{textcomp}
\usepackage{stfloats}
\usepackage{url}
\usepackage{verbatim}
\usepackage{graphicx}
\usepackage{cite}
\usepackage{xcolor}
\usepackage{bm}
\usepackage[]{units}
\usepackage{mathrsfs}
\usepackage{mathtools}
\usepackage{subfig}
\usepackage{wasysym} 
\usepackage{hyperref}
\usepackage{cleveref}
\Crefname{equation}{Eq.}{Eqs.}
\Crefname{figure}{Fig.}{Figs.}
\Crefname{section}{Sec.}{Secs.}
\crefname{appendix}{App.}{Apps.}

\DeclareMathSizes{10}{10}{6}{1}

\newcommand\mynotes[1]{{}}

\def\<{ \begin{bmatrix} }
\def\>{ \end{bmatrix} }

\renewcommand{\d}{{\mathrm{d}}} 			
\renewcommand{\c}{{\mathrm{c}}} 			

\renewcommand{\b}{{\bm{b}}}     			
\renewcommand{\phi}{\varphi} 		    
\newcommand{\current}{\bm{i}} 		    
\newcommand{\g}{{\bm{g}}}  				
\newcommand{\reg}{\textregistered\,}	
\newcommand{\SP}{\mathrm{SP}} 	 		
\newcommand{\m}{\tilde{\bm{m}}} 		
\newcommand{\n}{\bm{n}} 				
\newcommand{\emm}{\mathrm{m}}           
\newcommand{\K}{\bm{\mathrm{K}}}		
\newcommand{\M}{\bm{\mathrm{M}}}		
\newcommand{\V}{\bm{\mathrm{V}}}		
\newcommand{\U}{\bm{\mathrm{U}}}		
\newcommand{\R}{\bm{R}}		            
\newcommand{\x}{\bm{x}} 		 		
\newcommand{\Zero}{\bm{0}} 		 		
\newcommand{\I}{\bm{\mathrm{I}}} 		
\newcommand{\Act}{\bm{\mathcal{A}}} 	
\newcommand{\T}{\top} 			        
\newcommand{\p}{\bm{p}}                 
\newcommand{\torque}{\bm{\tau}}         
\newcommand{\force}{\bm{f}}             
\newcommand{\jacobi}{\bm{\mathrm{J}}}   
\newcommand{\mb}{|\tilde{\bm{m}}||\bm{b}|} 		
\newcommand{\skews}[1]{\mathrm{skew}\!\left\{#1\right\}}
\newcommand{\norm}[1]{\lVert #1 \rVert}
\newcommand{\abs}[1]{\lvert #1 \rvert}
\newcommand{\rank}[1]{\mathrm{rank}\,\{#1\}}
\newcommand{\range}[1]{\mathrm{range}\,\{#1\}}
\newcommand{\image}[1]{\mathrm{image}\,\{#1\}}    
\newcommand{\nulls}[1]{\mathrm{null}\,\{#1\}}
\newcommand{\spans}[1]{\mathrm{span}\,\{#1\}}

\newcommand{\Bprefix}[1]{\ensuremath{\prescript{}{\mathcal{B}}{#1}}}
\newcommand{\torqueB}{\Bprefix{\torque}}
\newcommand{\mB}{\Bprefix{\m}}
\newcommand{\bB}{\Bprefix{\b}} 
\newcommand{\aI}{{\ensuremath{(\mathrm{I})}}}  
\newcommand{\aII}{{\ensuremath{(\mathrm{II})}}} 

\usepackage{placeins}

\newlength{\figcapskip}
\newlength{\figimgcapskip}
\makeatletter
\def\@IEEEfigurecaptionsepspace{\vskip\abovecaptionskip\vskip\figimgcapskip\relax}
\makeatother

\makeatletter
\def\@IEEEBIOskipN{1.3\baselineskip}
\makeatother

\begin{document}

\title{Expanding the Workspace of Electromagnetic Navigation Systems Using Dynamic Feedback for Single- and Multi-agent Control}

\author{Jasan Zughaibi, Denis von Arx, Maurus Derungs, Florian Heemeyer, Luca A. Antonelli, Quentin Boehler, Michael Muehlebach, and Bradley J. Nelson
\thanks{This work was supported in part by the Swiss National Science Foundation under Grant IZLCZ0\_206033, in part by the Max Planck ETH Center for Learning Systems, and in part by the German Research Foundation. (\textit{Corresponding author: Jasan Zughaibi.})}

\thanks{Jasan Zughaibi, Denis von Arx, Maurus Derungs, Florian Heemeyer, Luca A. Antonelli, Quentin Boehler, and Brad Nelson are with the Multi-Scale Robotics Lab, ETH Zurich, 8092 Zurich, Switzerland (e-mail: \{zjasan; dvarx; mderung; fheemeyer; lucaan; qboehler;  bnelson\}@ethz.ch).}
\thanks{Michael Muehlebach is with the Learning and Dynamical Systems Group, Max Planck Institute for Intelligent Systems, 72076 Tübingen, Germany (e-mail: michael.muehlebach@tuebingen.mpg.de).}
\thanks{A video is available at \texttt{\url{https://youtu.be/PQeAKPL_iS0}}.}

}

\maketitle

\begin{abstract}
Electromagnetic navigation systems (eMNS) enable magnetically guided surgical procedures, but their effective workspace is often limited by power and thermal constraints. We show that system-level control significantly expands this workspace by reducing the required currents. Five key approaches enable this: (i) motion-centric torque/force objectives, (ii) energy-optimal current allocation, (iii) real-time pose estimation, (iv) feedback control, and (v) high-bandwidth eMNS components. Using these, we stabilize a 3D inverted pendulum on an eight-coil OctoMag eMNS with much lower currents (0.1–0.2 A vs. 8–14 A), replacing the field-centric field-alignment strategy with a motion-centric torque/force approach. We generalize to multi-agent control by simultaneously stabilizing two pendulums in a shared workspace, exploiting field nonlinearity and coil redundancy. A structured analysis compares the workspaces of both paradigms and current-allocation strategies. Cross-platform evaluation on the Navion eMNS confirms substantial workspace expansion, maintaining stable balancing up to 50 cm from the coils.
\end{abstract}



\section*{Nomenclature}
\addcontentsline{toc}{section}{Nomenclature}

\begin{IEEEdescription}[\IEEEsetlabelwidth{$\image{\cdot}xxx$}]
    \item[$\current$] Electrical current vector
    \item[$\b$] Magnetic field vector
    \item[$\g$] Magnetic gradient vector
    \item[$\m$] Magnetic dipole moment
    \item[$\p$] Position of the magnetic dipole
    \item[$\torque$] Magnetic torque vector
    \item[$\torque_\c$] Effective control torque vector
    \item[$\torque_d$] Torque from dipole-dipole interaction
    \item[$\torque_{\mathrm{net}}$] Net torque acting on magnetic volume
    \item[$\force$] Magnetic force vector
    \item[$\x$] State vector
    \item[$\Act$] Actuation matrix
    \item[$\Act_{\b}$] Field-actuation matrix
    \item[$\Act_{\g}$] Gradient-actuation matrix
    \item[$\M$] Wrench allocation matrix
    \item[$\M_\b$] Torque allocation matrix
    \item[$\M_\g$] Force allocation matrix
    \item[$\jacobi$] Jacobian matrix
    \item[$\K$] State feedback gain matrix
    \item[$\R$] Rotation matrix
    \item[$\I$] Identity matrix
    \item[$\Zero$] Zeros matrix
    
    \item[$\norm{\cdot}$] $\ell_2$ norm
    \item[$\skews{\cdot}$] Skew-symmetric matrix operator
    \item[$\rank{\cdot}$] Rank of a matrix
    \item[$\range{\cdot}$] Range of a matrix
    \item[$\image{\cdot}$] Image of a matrix
    \item[$\nulls{\cdot}$] Nullspace of a matrix
    \item[$\spans{\cdot}$] Span of a set of vectors
    \item[$(\cdot)_\SP$] Subscript denoting a setpoint or reference 
    \item[$\Bprefix{(\cdot)}$] Prefix denoting the body-fixed frame
\end{IEEEdescription}

\section{Introduction}
Magnetic navigation systems are rapidly emerging as a key technology in medical robotics, enabling breakthroughs from precision drug delivery to sophisticated endoscopic procedures \cite{zhao2022tele_mag, mesot2024teleoperated, schuerle2022microLiving}. These systems act on nanometer to centimeter scales and encompass both soft and hard magnetomagnetic materials \cite{cehmke2025remag, hertle2025covalently}. Dynamic magnetic fields can be generated in two principal ways: by robotic manipulators that position and orient permanent magnets in space~\cite{valdastri2019levitation}, or by electromagnetic navigation systems (eMNS) that produce fields using current-driven coils~\cite{simone2024navion}, \cite{liZhang2020review}. Recent\mynotes{change back to Our recent after doubleblind} work demonstrated high actuation bandwidth of an eMNS to effectively achieve complex dynamic tasks~\cite{zughaibi2024balancing, sydora2026learning}. In contrast, platforms that move permanent magnets are often credited with higher field strengths~\cite{kladko2024magnetosurgery}, but remain constrained by inertial mass, limiting the mechanical bandwidth of the system~\cite{hongsoo2020vascular}.

\begin{figure}
	\centering
	\includegraphics[scale=0.7,   trim = 4cm 10.5cm 3cm 8.5cm,
    clip ]{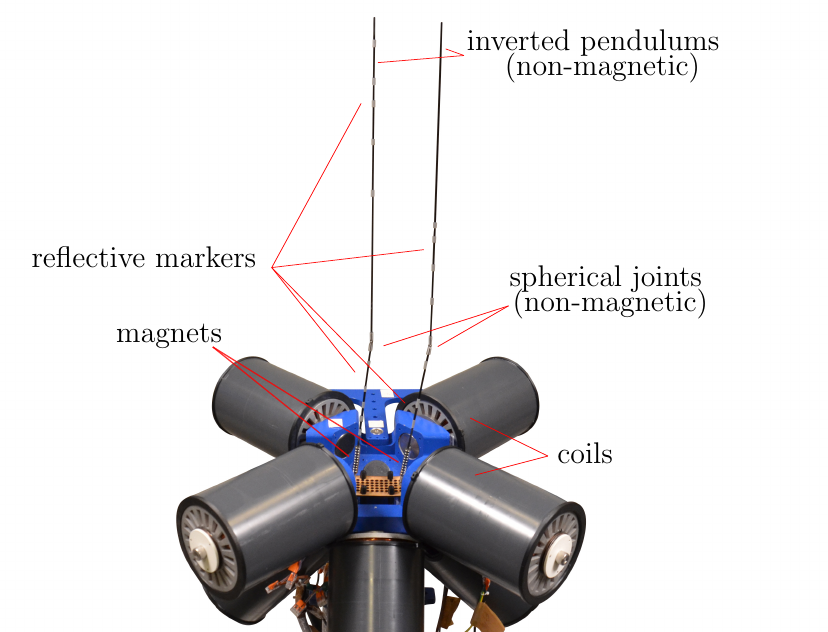} 
	\caption{We balance two 3D inverted pendulums simultaneously within the same magnetic workspace, leveraging the magnetic field created by the OctoMag eMNS. Because both pendulums are identical, independent actuation under a global field requires exploiting the nonlinearity of the magnetic field. This setup is used as an experimental platform to compare different strategies for multi-agent control. Each inverted pendulum system includes an arm driven by the external magnetic field and a non-magnetic pendulum. Balancing two inverted pendulums within the same magnetic workspace is challenging due to coupling effects not only between each coil and the permanent magnets, but also between the magnets themselves.}
	\label{fig:two_inv_pends}
\vspace{\figcapskip}
\end{figure}

A key challenge in electromagnetic navigation is generating magnetic fields that remain effective over clinically relevant distances to induce motion. This is inherently difficult because magnetic fields decay rapidly, typically with the inverse cube of distance. Moreover, there are infinitely many magnetic field configurations that produce the same torque on an object, making the problem fundamentally non-unique. As a result, the corresponding coil currents required to generate these fields can differ substantially, even though they induce identical motion. Consequently, the manner in which an eMNS is operated can significantly impact its effective range. Achieving high currents and power is technically demanding, primarily due to thermal limits, which complicates the design of large-scale systems. This limits the scalability of eMNS technologies and the scope of clinical applications \cite{simone2024navion}.

In this work, we comprehensively demonstrate that system-level control architecture is crucial for substantially reducing energy consumption in eMNSs, thereby directly expanding the effective magnetic workspace. We identify five architectural ingredients as key enablers: (i) adopting motion-centric torque and force objectives, (ii) allocating coil currents through energy-optimal control, (iii) integrating real-time pose estimation, (iv) employing feedback control, and (v) leveraging high-bandwidth eMNS platforms. 

Following the approach of~\cite{zughaibi2024balancing}, we evaluate our findings by dynamically stabilizing inverted pendulums mounted on magnetically actuated arms. This task is inherently challenging due to the system’s intrinsic instability, which demands continuous feedback control to reject disturbances and maintain equilibrium. Although not directly tied to clinical applications, this experiment serves as a valuable benchmark for exploring advanced dynamic magnetic control strategies.

We revisit the 3D inverted pendulum problem from~\cite{zughaibi2024balancing} using the OctoMag - an eight-coil eMNS - by replacing the original field-centric \textit{field-alignment} strategy with a motion-centric \textit{torque/force-based control} approach. Both paradigms are widely adopted in electromagnetic navigation: some rely on field-alignment, which exploits the natural tendency of a magnetic dipole to align with an external magnetic field~\cite{mahoney2016bAlignPermMag, stereotaxis2006align, catheter2011_eMNS_align, kummer2010octomag, filgueiras2013remote_align, salehizadeh2020assume_alignment, kumar2016indepIdent}, while others employ wrench-based control strategies~\cite{berkelman2013levitation, diller2014six_BF_TF}. We show that the same stabilization task from~\cite{zughaibi2024balancing} can be achieved with substantially lower coil currents (\unit[0.1–0.2]{A} instead of \unit[8–14]{A}) by adopting the torque/force-based approach. A key contribution of this work is a structured comparison of these two paradigms. We analyze the observed disparity in current requirements by comparing their electromagnetic workspaces, providing an in-depth theoretical analysis for different current allocation strategies (i.e., mappings from control inputs to coil currents), and discussing the respective advantages and limitations of each control method.

\begin{figure}
	\centering
	\setlength{\figimgcapskip}{-4mm}
	\includegraphics[scale=0.655,   trim = 4.4cm 10.5cm 3cm 7.5cm, clip ]{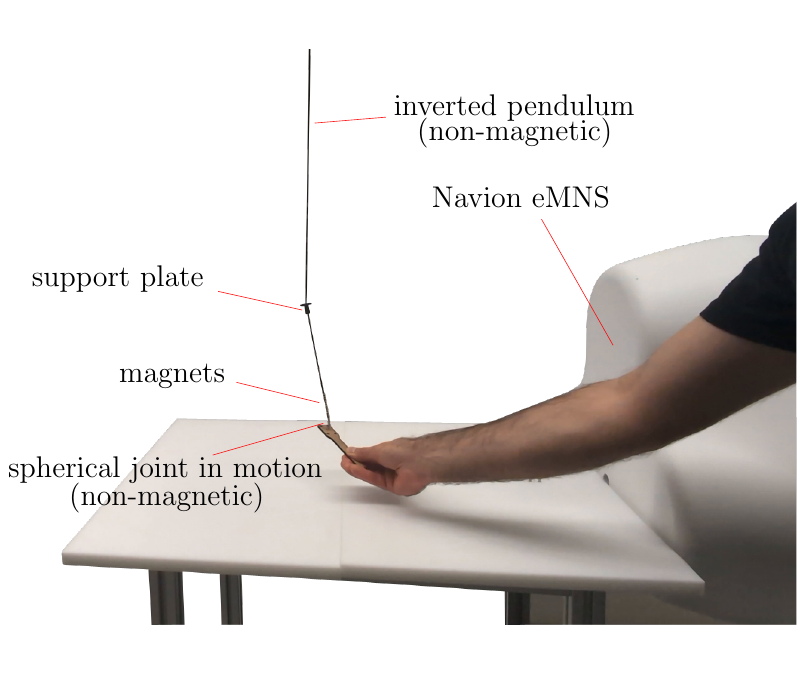}
	\vspace{-5mm}
    \caption{The figure highlights an inverted-pendulum balancing on a magnetically driven arm controlled by the clinically-ready Navion eMNS. While the arm is moved manually through space (see accompanying video), a closed-loop optimal-energy controller continuously transfers magnetic energy into the actuator’s motion with very high efficiency, so the pendulum remains stable even several coil diameters away from the coils. The feedback strategy therefore expands the eMNS’s effective electromagnetic workspace well beyond what open-loop control can achieve. In this setup, instead of connecting the actuator and inverted pendulum through a spherical joint, the pole tip is balanced directly on a (non-magnetic) 3D-printed support plate.}
	\label{fig:navion_inv_pend_moving_around}
\vspace{-3mm}
\end{figure}

We extend this systematic comparison of control paradigms to multi-agent scenarios, in which multiple magnetic agents are independently actuated by exploiting the nonlinearity of the magnetic field. To this end, we introduce a novel benchmark using the OctoMag eMNS: the simultaneous stabilization of two identical inverted pendulums within a shared magnetic workspace, as illustrated in \Cref{fig:two_inv_pends}. To appreciate the difficulty of this scenario, imagine a person attempting to balance two poles simultaneously, one in each hand. Such capabilities could potentially open up new clinical applications, where multiple magnetic instruments are controlled independently within dynamic environments, for example, to compensate for physiological motion. In these scenarios, the system-level design principles identified earlier remain essential, as they help mitigate the strong coupling effects that naturally arise between the coils and the magnetic objects. To our knowledge, this is the first demonstration of simultaneous control of two inverted pendulums in a shared magnetic environment.

Finally, we employ the Navion eMNS, a system designed for medical applications, to stabilize the inverted pendulum (see \Cref{fig:navion_inv_pend_moving_around}). This experiment highlights the significant workspace expansion enabled by our approach, with the pendulum stabilized at a distance of \unit[50]{cm} from the coils. It also serves to demonstrate the cross-platform applicability of our control strategy.

These experiments, ranging from single- to multi-agent control, and across different hardware platforms, underscore the importance of system-level design choices in enabling advanced magnetic manipulation. To further emphasize this point, we deliberately adopt simple linear feedback strategies, such as Linear Quadratic Regulators (LQR) augmented with integral action. Rather than optimizing the feedback algorithms themselves, our goal is to isolate and demonstrate the impact of architectural decisions on overall system performance. We adopt a modeling approach based on linear dynamic models, relying on the assumption that high-bandwidth feedback can mitigate model uncertainties. 

A critical enabler of our methods is real-time state information, and we emphasize that clinical translation depends on the development of in-vivo state estimation techniques that are non-invasive and safe for biological tissue. Intraoperative modalities, such as X-ray fluoroscopy, remain essential for high-level path planning as they provide the critical context regarding the tool's relative pose with respect to the patient's anatomy and organs. However, to minimize patient exposure to ionizing radiation, these anatomical imaging techniques are typically only used intermittently. Although our approach demonstrates robustness to low sampling rates and latency, the proposed control strategy strictly requires continuous low-level localization (i.e., tracking the object's pose relative to the eMNS coordinate frame). Commercial electromagnetic tracking solutions (e.g., NDI\textregistered, Quadrant Scientific\textregistered) are widely used in surgical navigation, but their compatibility with the dynamic actuation fields of an eMNS remains largely unverified. Recent research addresses this gap, demonstrating that high-frequency state estimation can be performed safely and concurrently within an eMNS actuation field via the superposition of high-frequency fields and the integration of Hall-effect sensors~\cite{denis2024simulLocAct} or inductive pick-up coils~\cite{denis2025pickupcoil}.

The paper is organized as follows. \Cref{sec:Related Works} discusses related works, with an emphasis on multi-agent control. \Cref{sec:Experimental Platform} details the experimental platforms, i.e., the eMNS hardware, the inverted-pendulum setups, the system dynamics common to both actuation paradigms, and the feedback control algorithms. \Cref{sec:Single-Task_Control} compares field-alignment and torque/force-based paradigms for single-agent scenarios, including electromagnetic-workspace characterization and a theoretical analysis of different current allocation strategies. \Cref{sec:Multi-Task_Control} extends this comparison to multi-agent settings. Finally, \Cref{sec:Discussion} summarizes the findings and discusses implications and future directions.

\section{Related Work}
\label{sec:Related Works}

A comprehensive review of dynamical-systems approaches to magnetic navigation is given in \cite{zughaibi2024balancing}. We therefore focus on related work in multi-agent magnetic navigation and inverted pendulum systems comparable to our multi-agent platform.

The control of multiple magnetic agents, including swarms, is a well-established topic in microrobotics. A comprehensive overview of the field is provided in \cite{review2021microbSwarms}. Various approaches exist for controlling multiple magnetic objects in space. Some methods exploit asymmetries in the object properties - such as differences in geometry or magnetization - to induce distinct responses under a global control input \cite{diller2013indep3D}, \cite{aaronBecker2013}. Of interest for our work, is the area of leveraging of inhomogeneity of magnetic fields, by explicitly shaping spatially varying fields to control multiple objects at distinct locations. Notably, the two inverted pendulums in our study are identical in both geometry and magnetization, meaning our method does not rely on heterogeneity to enable independent control. This introduces an additional challenge, as common techniques such as time-scale separation cannot be exploited to decouple the system dynamics. Naturally, our methods can also be applied to non-identical objects.

Previous work has demonstrated the control of identical magnetic robots. For example, \cite{kumar2016indepIdent} employs PID control running at \unit[10]{Hz} to independently position two identical, millimeter-scale magnetic robots while neglecting rotational dynamics, assuming stable field alignment. Similarly, \cite{misra2019eMNSindentNonIdent} controls both identical and non-identical soft-magnetic microrobots using a nine-coil system, leveraging magnetic field inhomogeneity. In their work, control currents are computed by solving a nonlinear optimization problem to minimize power consumption and heat dissipation. The system operates with PID controllers running at \unit[25]{Hz}.

Multi-agent magnetic control does not necessarily imply full independence of objects. For example, in \cite{misra2021multiPointCR}, a magnetic continuum robot incorporating two discrete magnets within a soft material is controlled using a PI control scheme. By exploiting field inhomogeneity, the system is used to control desired shape deformations based on a quasi-static Cosserat rod model, neglecting dynamic effects. Due to computational complexity, the control loop operates at \unit[2]{Hz}.

Although direct comparisons to existing multi-agent inverted pendulum studies are limited, insights can be drawn from research on underactuated systems. Parallel-type double inverted pendulums, as studied in \cite{lu2007parallelInvPend}, feature two inverted pendulums mounted on a single cart, designed with differing geometries to introduce time-scale separation to achieve controllability. As already mentioned further above, our work does not rely on such assumptions, and both pendulums are identical. Nonetheless, both setups share the fact that a single actuation medium controls multiple pendulums, leading to coupling effects similar to those in parallel-type pendulums. A related example is \cite{hofer2023owc}, where a single reaction wheel stabilizes a three-dimensional inverted pendulum, with inertia ratios computed for an optimal time-scale separation. It is important to note that our two-pendulum system is not underactuated from a control-theoretic standpoint, as the OctoMag eMNS employs eight coils.

\section{Experimental Platforms, Dynamics, and Control}
\label{sec:Experimental Platform}

This section details the experimental setups used throughout the study for evaluating the control paradigms in both single- and multi-agent scenarios.
We also describe the two eMNS employed, and the system dynamics common to both control approaches. Finally, the two feedback control algorithms used for stabilization are presented. 

\subsection{Inverted Pendulum Platforms}

\begin{figure}
	\centering
	\includegraphics[scale=0.87, trim = 7.0cm 16cm 4.5cm 3.5cm, clip]{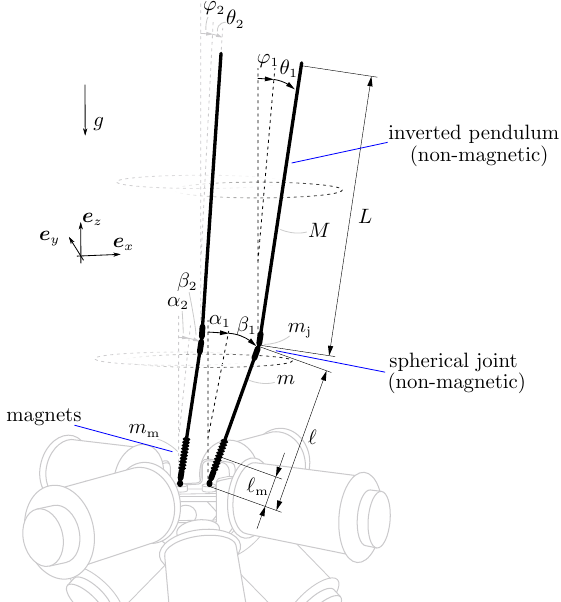} 
	\caption{Schematic of orientation conventions for the multi-agent platform. Each pendulum–actuator unit has four DOFs: actuator tilts $(\alpha_i,\beta_i)$ and pendulum tilts $(\varphi_i,\theta_i)$, with $i\in\{1,2\}$ for left/right. The pendulums are separated by $\approx\unit[6.5]{cm}$. Masses: actuator rod $m$, pendulum rod $M$, spherical joint $m_\mathrm{j}$, magnets $m_\mathrm{m}$. Lengths: actuator $\ell$, pendulum $L$, and $\ell_\mathrm{m}$ (distance from magnet centers to the pivot). Figure adapted from \cite{zughaibi2024balancing}.}
	\label{fig:two_inv_pends_sketch}
\vspace{\figcapskip}
\end{figure}

We employ a set of custom built inverted pendulum platforms as benchmark systems for evaluating the proposed control paradigms in both single- and multi-agent scenarios. The platforms differ in complexity and task configuration but share a common mechanical structure. Each pendulum assembly consists of two rigid, carbon-fiber reinforced thermoplastic rods: the \textit{actuator} (lower rod), which is magnetically actuated and carries ten axially magnetized permanent magnets (Magnetkontor\reg~R-08-03-04-N3-N, NdFeB N45, \diameter8 (3) × 4 mm), and the \textit{pendulum} (upper rod), which is non-magnetic. We use three primary platform variants:

\begin{itemize}
    \item \textit{Single-task platform (1$\times$3D):} A single 3D inverted pendulum system, previously described in \cite{zughaibi2024balancing}, serves as the baseline. The actuator is mounted to a base plate through a non-magnetic, spherical, universal joint (303 stainless steel, McMaster-Carr\reg~60075K75), allowing free rotation in two angular directions, parameterized by $(\alpha, \beta)$. The pendulum is connected to the actuator through an identical joint, yielding a second pair of angular degrees of freedom $(\varphi, \theta)$. To better illustrate the pendulum's unconstrained, free-falling behavior, we also employ a joint-less variation - shown in \Cref{fig:navion_inv_pend_moving_around} - in which a sharpened pendulum rod is balanced directly on a flat plate without mechanical connection. This configuration is used with the Navion eMNS for large-workspace experiments. 
    \item \textit{Multi-agent planar platform (2$\times$2D):} This configuration consists of two independent planar inverted pendulums, each constrained to move within a 2D plane. Each assembly is connected to the base and to the pendulum through 3D-printed plastic joints incorporating non-magnetic plastic bearings (Igus\reg BB-604-B180-30-GL). The left system is described by $(\alpha_1, \varphi_1)$ and the right by $(\alpha_2, \varphi_2)$. 
    \item \textit{Multi-agent platform (2$\times$3D):} This platform consists of two fully independent 3D inverted pendulum systems, each constructed as in the single-task setup and mounted on a shared base. The actuator orientations are parameterized by $(\alpha_1, \beta_1)$ and $(\alpha_2, \beta_2)$, and the corresponding pendulum orientations by $(\varphi_1, \theta_1)$ and $(\varphi_2, \theta_2)$, respectively.
\end{itemize}
The angle definitions for the 2$\times$3D case are illustrated in \Cref{fig:two_inv_pends_sketch}, and apply analogously to the 2$\times$2D and 
1$\times$3D configurations. Reflective marker stripes are placed on each actuator and each pendulum rod, enabling a motion capture system to record all angles in real-time; in practice, its update rate is aligned with the driver communication frequency of each eMNS platform, as reported below.

\subsection{Electromagnetic Navigation Systems}
This section provides a brief overview of the electromagnetic navigation systems (eMNSs) used in this study and highlights their key characteristics. A crucial property for the control strategies presented is the actuation bandwidth. 

The bandwidth of an eMNS refers to the frequency at \unit[-3]{dB} gain reduction beyond which the current controller can no longer follow its reference. This bandwidth must be sufficiently high to ensure that the magnetic torques and forces commanded by higher-level feedback controllers closely match the actual torques and forces applied to the object. Lower bandwidths introduce lag, increasing the effective time constant and settling time of the torque/force dynamics. The required bandwidth depends on the target application and in particular on the frequency range in which accurate tracking and effective disturbance rejection are necessary. From a control perspective, the system bandwidth should be at least several times or ideally an order of magnitude greater than the highest frequency of interest, such as a reference signal or disturbance. For example, in cardiac ablation procedures, a catheter inside the beating heart can experience disturbances near \unit[3]{Hz}, corresponding to a \unit[180]{bpm} heart rate, which implies that the actuation bandwidths for such scenarios must exceed this by a significant margin.

This bandwidth depends on the hardware, particularly the current drivers, coil parameters, but also the applied current amplitude. All reported bandwidths are identified at an amplitude of \unit[5]{A}. 

\paragraph{OctoMag}
The OctoMag is a research-oriented eMNS consisting of eight independently driven coils (see \Cref{fig:two_inv_pends_sketch}), as described in \cite{kummer2010octomag}. It is controlled with a current input vector $\current \in \mathbb{R}^8$ using custom-built driver electronics. The closed-loop current control bandwidth was identified in \cite{zughaibi2024balancing} as \unit[26.4]{Hz}. Each coil is limited to a maximum current of \unit[16]{A} during operation. Current setpoints are updated at a control frequency of \unit[200]{Hz}.

\paragraph{Navion}
The Navion is a preclinical eMNS comprising three parallel coils in a triangular arrangement~\cite{simone2024navion}, as illustrated in \Cref{fig:navion_inv_pend_moving_around}, with $\current \in \mathbb{R}^3$. It is described in detail in \cite{simone2024navion} and uses commercial driver electronics (Gold Drum HV, Elmo Motion Control\reg). Using the system identification method from \cite{zughaibi2024balancing}, the electrical bandwidth is identified to \unit[24.5]{Hz}. Although the hardware supports coil currents up to \unit[45]{A}, we conservatively restrict them to \unit[25]{A} per coil for safety. Current setpoints are sent to the driver at a control frequency of \unit[125]{Hz}.
 
\subsection{Relation between Magnetic Fields and Currents}
We assume that the magnetic field $\b$ and its gradient $\g \coloneqq \big(\begin{array}{lllll}
	\frac{\partial b_x}{\partial x} & \frac{\partial b_x}{\partial y} & \frac{\partial b_x}{\partial z} & \frac{\partial b_y}{\partial y} & \frac{\partial b_y}{\partial z}
\end{array}\big)^\T$ depend linearly on the electrical currents $\current \in \mathbb{R}^n$, $n\in \{ 3, 8 \}$. This relationship is commonly expressed as:
\begin{align}
    \< \b \\ \g \> = \< \Act_{\b}(\p) \\ \Act_{\g}(\p) \> \bm{i} = \Act(\p) \current.
\end{align}
Here, $\Act(\p)$ denotes the position-dependent actuation matrix, which we update in real time based on the magnetic dipole's position $\p \in \mathbb{R}^3$, as measured by the motion capture system. This matrix is analytically derived from the nonlinear multipole-expansion model, combined with a system-specific calibration procedure, as described in \cite{petruska2017multipole, bernardes2026structured}. To ensure accurate spatial alignment during this process, calibration is performed using a 4x4x4 array of Hall-effect sensors tracked directly by the motion-capture system, which ensures that the magnetic center is coincident with the inertial tracking frame used during real-time evaluation\footnote{Field data was collected at 64 positions for the Navion and 320 for the Octomag, yielding an error of $<$\unit[0.2]{mT} on the test set (a 20\% subset of the data) for the specific calibration models used herein. However, a more comprehensive spatial analysis \cite{bernardes2026structured} bounds realistic, workspace-wide root-mean-square (RMS) errors to $<$\unit[0.5]{mT} for the Octomag and $<$\unit[3]{mT} for the Navion.}.
We refer to $\Act_{\b} \in \mathbb{R}^{3 \times n}$ as the \textit{field-actuation matrix} and to $\Act_{\g} \in \mathbb{R}^{5 \times n}$ as the \textit{gradient-actuation matrix}. Although the magnetic field gradient $\nabla \b$ is a $3 \times 3$ matrix, it is fully described by only five independent components due to Maxwell’s equations under the assumption of a static field with no free currents. Specifically, the constraints $\nabla \cdot \b = 0$ and $\nabla \times \b = 0$ reduce the degrees of freedom from nine to five.

\subsection{Common Dynamics}
Central to our approach is a perspective rooted in dynamical systems. While some aspects of the system dynamics are shared across the two paradigms, key differences arise due to the nature of the control inputs - whether one employs the field-alignment technique or the torque/force-based control paradigm. In particular, different representations of the actuator dynamics must be considered, while the equation for the pendulum remains unchanged.

We derive the equations of motion using the Lagrangian formalism. The methodology for obtaining the equations of motion for the three-dimensional (3D) inverted pendulum is presented in detail in \cite{zughaibi2024balancing}. In the present work, we revisit the key steps to highlight the differences that arise when choosing distinct control inputs. In \cite{zughaibi2024balancing}, we showed that the 3D pendulum dynamics may be adequately approximated by two decoupled two-dimensional (2D) pendulums for the purposes of controller design. This approximation allows us to use the same 2D equations of motion for both the 1$\times$3D setup, as well as the 2x2D and the 2x3D configurations employed in our multi-agent experiments. For brevity, we provide the derivation only for the $(\alpha, \varphi)$ plane, noting that analogous equations hold for the $(\beta, \theta)$ plane.

Let $T$ and $U$ denote the kinetic and potential energies of the mechanical system:
\begin{align*}
	T &=\frac{1}{2}\left(J+M \ell^2\right) \dot{\alpha}^2+\frac{1}{8} M L^2 \dot{\varphi}^2+\frac{1}{2} M \ell L \dot{\alpha} \dot{\varphi} \cos (\alpha-\varphi) \\
	U &= (\eta+M \ell) g \cos \alpha+M g \frac{L}{2} \cos \varphi,
\end{align*}
where $\eta$ (in $\mathrm{kg\,m}$) and $J$ (in $\mathrm{kg\,m^2}$) are the first and second moments of mass of the actuator, with full expressions given in \cite{zughaibi2024balancing}. Magnetic potential energy is excluded from $U$ and is only introduced in the field-alignment formulation, as discussed in the next section. The Lagrangian of the mechanical system is defined as $\mathscr{L} \coloneqq T - U$. While the actuator dynamics vary between control strategies, the pendulum’s equation of motion remains unchanged. The differential equation governing the generalized coordinate $\varphi$ is obtained from Lagrange's equation
\begin{align}
    \frac{\d }{\d t} \biggl( \frac{\partial \mathscr{L}}{\partial \dot{\varphi}}\biggr) - \frac{\partial \mathscr{L}}{\partial \varphi} = 0
\end{align}
which, upon linearization, yields
\begin{align*}
    \frac{1}{4} M L^2 \ddot{\varphi}+\frac{1}{2} M \ell L \ddot{\alpha}- M g \frac{L}{2} \varphi=0
\end{align*}
The more critical equation for this work is the one governing the generalized coordinate $\alpha$, as its exact form depends on the selected control input. The specific details are provided in the respective sections (see \Cref{sec:Field_Alignement_Control} and \Cref{sec:Torque/Force-based Control}), illustrating how each control approach modifies the system equations and thus influences the controller design.

\subsection{Feedback Control}
To stabilize the system, we employ a Linear Quadratic Regulator with Integral action (LQRI) applied in a fully decoupled manner, which is a standard state-feedback control policy. The control law for pendulum system $i \in \{ 1, 2\}$ at timestep $k$ takes the form:
\begin{align}
	\< \xi_i[k] \\ \eta_i[k] \> = \< \K_\alpha & \Zero \\ \Zero &  \K_\beta \> \< \x_{\alpha, \SP, i}[k] - \x_{\alpha, i}[k] \\ \x_{\beta, \SP, i}[k] - \x_{\beta, i}[k] \>, \label{eq:control_policy}
\end{align}
where the state vectors are defined as 
\begin{align*}
    \x_{\alpha, i} \coloneqq \< \alpha_i & \varphi_i & \dot{\alpha}_i & \dot{\varphi}_i \>^\T, \qquad \x_{\beta, i} \coloneqq \< \beta_i & \theta_i & \dot{\beta}_i & \dot{\theta}_i  \>^\T
\end{align*}
\noindent and setpoints are denoted by~$\x_{\alpha, \SP, i}$ and ~$\x_{\beta, \SP, i}$. Here $(\xi_i, \eta_i)$ corresponds to either the field angles $(u_{\alpha, i}, u_{\beta, i})$ for the field-alignment controller, or to the control torques $(\tau_{\mathrm{c}, \alpha, i}, \tau_{\mathrm{c}, \beta, i})$ for the torque-based controller (see \Cref{fig:field_torque_blockdiagram}). For the single-agent case, we omit the index $i$ for notational simplicity. 

The feedback gains $\K_\alpha$ and $\K_\beta$ are computed by solving the associated discrete-time algebraic Riccati equations based on the corresponding linearized system dynamics, which differ depending on the chosen control input formulation. Full state feedback is achieved by computing angular velocity terms using finite differences. 

To eliminate steady-state error, we incorporate decoupled integral controllers, as illustrated in the block diagram in \Cref{fig:field_torque_blockdiagram}. The integrator for the $\alpha_i$-loop is defined as
\begin{align}
\xi_{\mathrm{I}, i} = k_\mathrm{I} \int (\alpha_{\SP, i} - \alpha_i) \mathrm{d}t,
\end{align}
and analogously for $(\eta_{\mathrm{I}, i}, \beta_i)$. For stabilization experiments, the integrators are typically enabled only until the offsets are removed and then switched off. For  tracking setpoint trajectories, they remain active throughout the experiment. 

Although the controller is designed in a decoupled structure, it exhibits sufficient robustness by rejecting coupling effects as external disturbances. All controllers are implemented in Python within the Robot Operating System (ROS) framework. The computational cost of running the feedback algorithms is negligible, which allows them to be deployed at high sampling rates. Since control updates are applied at each time step, a simple linear dynamical model is often sufficiently descriptive, as long as it captures the dominant system behavior and stability-critical dynamics. In many cases, frequent feedback with a moderately accurate model can be more effective than infrequent updates based on a highly precise model. A practical guideline for electromagnetic navigation is to use sampling rates of at least \unit[20]{Hz}, based on frequency-domain analysis in clinical settings reported in \cite{zughaibi2024balancing}. Similarly, communication delays should be kept below \unit[40]{ms} to avoid destabilizing phase lag (for a detailed analysis refer to \Cref{sec:Influence of Sampling Rate and Delays}).

\section{Single-Agent Control}
\label{sec:Single-Task_Control}
This section compares field-alignment and torque/force-based control strategies in the context of single-agent stabilization. We highlight how the control inputs differ in their dynamic formulation depending on the selected strategy, provide a workspace analysis, and a theoretical discussion of allocation strategies. Experimental results are presented for both the Navion and OctoMag eMNS platforms. 

\subsection{Field-Alignment Control}
\label{sec:Field_Alignement_Control}

For the \emph{field-alignment control} paradigm, either a feedback controller or a human operator command a desired magnetic field vector $\b$ to control the orientation of a magnetic agent. 

\subsubsection{Dynamics and Stabilization without Digital Feedback}

When operated manually without digital feedback, this approach is often considered for its intuitiveness, as magnetic dipole $\m$ naturally aligns with an external magnetic field $\b$. It is widely used in practice - for instance, when a human operator steers the field direction via joystick - making open-loop field control a natural choice in many applications \cite{mahoney2016bAlignPermMag, stereotaxis2006align, catheter2011_eMNS_align, kummer2010octomag, filgueiras2013remote_align}.
We seek to analyze this field-alignment behavior from a dynamical systems viewpoint. To this end, we adopt an energy-based formulation and derive the resulting dynamics. The potential energy of a magnetic dipole $\m$ in an external magnetic field $\b$ is given by $U_\mathrm{m} = -\m \cdot \b.$

Consider a two-dimensional (2D) plane, $\m$ and $\boldsymbol{b}$ can be parametrized by the angles $\alpha$ and $u_\alpha$ as:
\begin{align}
    \m = |\m| \< \sin \alpha \\ \cos \alpha \> \qquad \b = |\b| \< \sin u_\alpha \\ \cos u_\alpha \>. \label{eq:dipole_field_parametrization}
\end{align}
where $u_\alpha$ is the operator input, effectively describing the orientation of the magnetic field $\b$. Substituting these expressions into the potential energy yields  $
U_{\mathrm{m}}= - \mb  \cos(u_\alpha - \alpha).
$

To obtain the dynamical equation governing $\alpha$, we include this magnetic potential energy in the Lagrange formalism. The equation of motion for coordinate $\alpha$ is then given by 
\begin{align}
    \frac{\d }{\d t} \biggl( \frac{\partial \mathscr{L}}{\partial \dot{\alpha}}\biggr) - \frac{\partial \mathscr{L}}{\partial \alpha} + \frac{\partial U_\mathrm{m}}{\partial \alpha} = 0. \label{eq:lagrange_eq_field_based}
\end{align}
 For simplicity, consider only the dynamics of the actuator itself (i.e., no attached pendulum, so $M=0$). The resulting nonlinear equation of motion becomes
\begin{align}
J \ddot{\alpha}-\eta g \sin \alpha= |\m| |\b| \sin(u_\alpha - \alpha). \label{eq:nonlinear_eom_actuator}
\end{align}
\begin{figure}
	\begin{flushleft}
        \hspace{2mm}
		\includegraphics[trim={3.4cm 14.3cm 1.5cm 11.8cm},clip, scale=0.50]{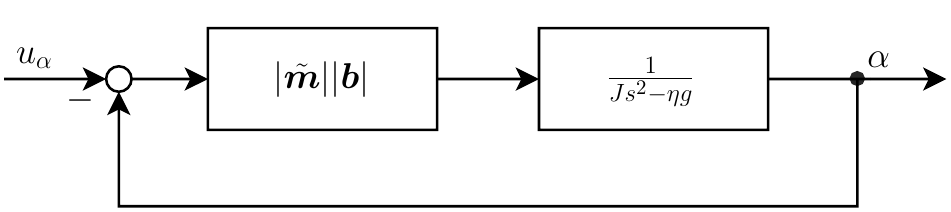} 
	\end{flushleft}
	\caption{Block diagram illustrating how open-loop field-alignment can be interpreted as a magnetic field acting as a proportional controller. The diagram is mathematically equivalent to the linearized dynamics in \eqref{eq:line_eom_actuator_P_ctrl} and \eqref{eq:line_eom_actuator_field_based}. In this formulation, the input $u_\alpha$ parameterizes the orientation of the magnetic field vector. Due to the proportional-like nature of the magnetic torque, the dipole naturally tends to align with the applied field direction. Open-loop field-alignment is commonly used in electromagnetic navigation due to its intuitive nature. However, viewing it from a control systems perspective clarifies both its capabilities and inherent limitations. }
	\label{fig:blockdiagram_field_P_ctrl}
\vspace{\figcapskip}
\end{figure}
\noindent Note, that the right-hand side corresponds to the magnetic torque expression $\m \times \b$ (see Sec.~\ref{sec:Torque/Force-based Control}). Under field alignment, the dipole remains nearly aligned with the magnetic field, resulting in a small angular difference between the dipole and field vectors. Accordingly, we apply a small-angle approximation to \eqref{eq:nonlinear_eom_actuator} to obtain the linearized dynamics.
\begin{align}
J \ddot{\alpha} -\eta g \alpha &= |\m| |\b| (u_\alpha - \alpha) \label{eq:line_eom_actuator_P_ctrl} \\
\Leftrightarrow J \ddot{\alpha} + ( -\eta g + |\m| |\b|) \alpha &= |\m| |\b| u_\alpha \label{eq:line_eom_actuator_field_based}
\end{align}
In \eqref{eq:line_eom_actuator_P_ctrl}, the term $(u_\alpha - \alpha)$ can be interpreted as a control error modulated by a proportional gain of $|\m||\b|$. These equations correspond directly to the closed-loop structure illustrated in \Cref{fig:blockdiagram_field_P_ctrl}. This offers a useful control-theoretic interpretation: if a human operator or feedforward controller continuously adjusts $u_\alpha$ to the desired dipole orientation, the magnetic field itself acts as a proportional feedback element that produces corrective torque. This viewpoint yields several key insights: 
\begin{itemize}
    \item \textit{Steady-State Offset:} From a control systems perspective it can be shown that purely proportional control implies a non-zero steady-state error under constant disturbances (e.g. caused by gradients). The larger $|\m||\b|$, the smaller the steady-state offset.
    \item \textit{Stability without Digital Feedback:} 
    If the magnetic field strength $|\b|$ or the magnetic dipole moment $|\m|$ is sufficiently large such that $-\eta g + |\m||\b| > 0$, the system becomes open-loop stable. In this case, the actuator aligns itself with the field even in the absence of explicit feedback control. This is particularly useful in applications where direct sensing is unavailable. Note, however, that this can be challenging for certain eMNS configurations, as gradients may introduce destabilizing forces. For instance, in the Navion system, an inherent gradient favors the lateral coil configuration, making it difficult to maintain a catheter or dipole in the upright orientation.
    \item \textit{Incompatibility with some Dynamical Systems:} While intuitive and suitable in a quasi-static domain, field-alignment may be incompatible with certain dynamic systems (e.g. the actuator or a catheter). In such cases, purely proportional magnetic torque can excite oscillatory modes.
\end{itemize}
This perspective is particularly relevant in clinical contexts, where magnetic systems are often operated without digital feedback. 

It must be noted that the analysis presented here assumes a homogeneous (gradient-free) field. Incorporating gradients directly into the field-alignment paradigm introduces more complexity, as detailed in App.~\ref{sec:appendix_gradient_incorp}.

\subsubsection{Field-Alignment with Digital Feedback}

\begin{figure*}
	\centering
	\includegraphics[scale=0.95,   trim = 6.0cm 8.5cm 6.3cm 2.5cm, clip ]{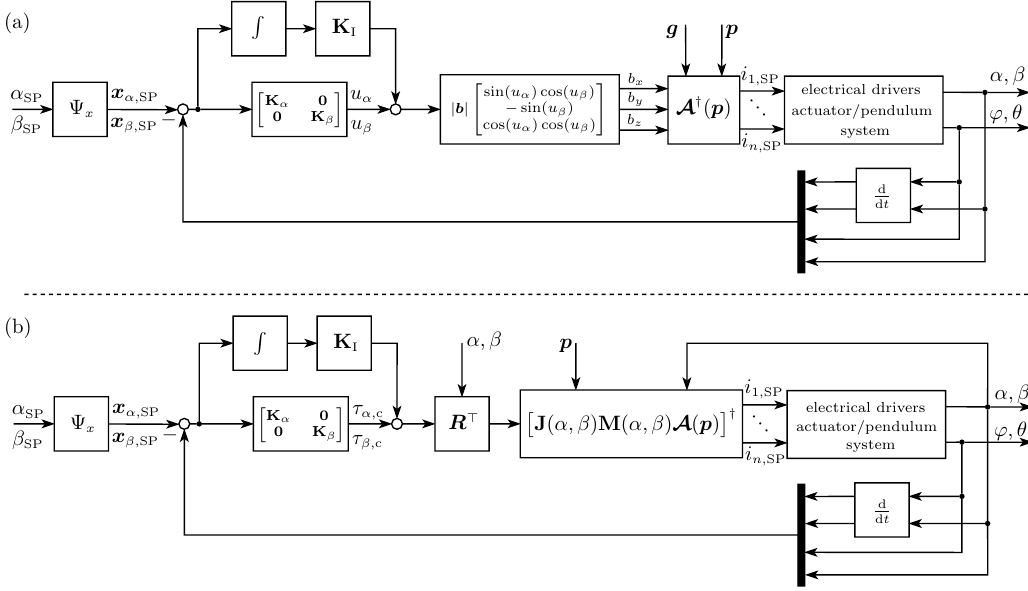} 
	\caption{Closed-loop control architecture for field-alignment control a) and torque-based control with minimum-energy allocation b), used for stabilizing the single 3D inverted pendulum. The key difference lies in the allocation strategy: while the field-alignment approach does not account for the current orientation of the magnetic dipole, the torque-based method explicitly requires the dipole's orientation in order to compute the corresponding currents. The setpoints $\alpha_\SP$ and $\beta_\SP$ are mapped to the corresponding state vectors using $\mathrm{\Psi}_{x, \alpha}: \alpha_\SP \mapsto \< \alpha_\SP & 0 & \dot{\alpha}_\SP & 0 \>^\T$, and analogously for $\beta_\SP$. Integral control is applied only to the angles $\alpha$ and $\beta$, i.e. using the gain matrix $\K_{\mathrm{I}} = k_\mathrm{I} \I^{2 \times 2} \otimes \operatorname{diag}\{\< 1 & 0 & 0 & 0 \>\}$. For the field-alignment approach, the gradients are set to $\g = 0$. In multi-task applications, the overall architecture remains decoupled, though the allocation strategy is adjusted accordingly. Note that the closed-loop field-alignment architecture in a) is presented strictly as a comparative benchmark, since the availability of state feedback renders it practically obsolete in favor of the more efficient torque-based approach.}
	\label{fig:field_torque_blockdiagram}
\vspace{\figcapskip}
\end{figure*}

The previously stated limitations in open-loop operation - such as steady-state error and active control of oscillations - can in principle be addressed through feedback control. Furthermore, to stabilize dynamic systems such as the inverted pendulum, digital feedback control is essential. However, it must be emphasized that if real-time state measurements are available to close the loop, employing field-alignment control becomes practically obsolete. Because real-time state estimation is the main prerequisite for implementing the highly energy-efficient torque/force-based control (detailed in Sec.~\ref{sec:Torque/Force-based Control}), there is no practical advantage to retaining the field-alignment paradigm once sensing is introduced. Its true utility remains strictly in sensorless, open-loop applications. Consequently, the closed-loop field-alignment architecture, illustrated in \Cref{fig:field_torque_blockdiagram} a), is presented here primarily as a comparative baseline to demonstrate its inherent inefficiencies. While the actuator dynamics remain inherently stable due to the magnetic field’s alignment behavior described earlier, the pendulum is dynamically unstable and thus requires active feedback for stabilization. The coil currents generated to maintain stability are shown in \Cref{fig:current_comparison_single_task} a), requiring currents in the range of \unit[8-14]{A}. Additional data is available in \cite{zughaibi2024balancing}, where comprehensive results are presented for stabilizing the 3D inverted pendulum using field-alignment control.

\subsubsection{Allocation Strategy} 

Given the outputs $u_\alpha, u_\beta$ from either a feedback controller or a human operator, the desired currents for the OctoMag eMNS are obtained by 
\begin{align}
	\current_\SP = \Act^\dagger(\p) \< \b_\SP\\ \Zero \>,\hspace{-2mm}\qquad\hspace{-2mm} \b_\SP = |\bm{b}| \< \sin(u_\alpha)\cos(u_\beta) \\  -\sin(u_\beta) \\ \cos(u_\alpha)\cos(u_\beta) \>, \label{eq:field_based_allocation}
\end{align}
where $|\b|$ is the user-defined field magnitude. The allocation is identical in both open- and closed-loop operation. The use of the pseudoinverse $\Act^\dagger \in \mathbb{R}^8$ ensures a least-squares solution - i.e., the current vector that minimizes the squared error in achieving the specified \textit{magnetic task} (desired magnetic field and zero gradient, in this case). However, note that this is not equivalent to minimizing the electrical energy required for a given \textit{motion task}.

\subsection{Torque/Force-based Control}
\label{sec:Torque/Force-based Control}

Here, we analyze the torque/force-based control methodology, where the feedback controller directly computes a desired torque $\torque$ and/or force $\force$, as shown in the block diagram in \Cref{fig:field_torque_blockdiagram}. Unlike the field-alignment approach discussed in the previous section, this method makes \emph{no assumptions} about the orientation of the magnetic field relative to the magnetic dipole or the field's spatial homogeneity. Instead, both the field vector and any induced gradient emerges from the allocation process, rather than being imposed a priori through simplifying assumptions such as the small-angle approximation used in field alignment.

\subsubsection{Dynamics and the Importance of Feedback Control}

Unlike the field-alignment approach discussed in the previous section, the Lagrangian for the torque/force-based method does not include magnetic potential energy. Instead, the control input enters the dynamics as non-conservative generalized force:
\begin{align}
    \frac{\d }{\d t} \biggl( \frac{\partial \mathscr{L}}{\partial \dot{\alpha}}\biggr) - \frac{\partial \mathscr{L}}{\partial \alpha}  = \tau_{\c, \alpha}. \label{eq:lagrange_eq_torque_force}
\end{align}
For the special case where no pendulum is attached and $M = 0$, the equation of motion simplifies to:
\begin{align}
    J \ddot{\alpha} -\eta g \sin \alpha= \tau_{\c, \alpha}. 
    \label{eq:actuator_eom_no_pend_torque_force}
\end{align}
This describes the dynamics of the actuator alone, or similarly, a free magnetic dipole. Notably, the dynamics are unstable, which renders direct torque-based feedforward control impractical for a human operator, as unstable systems require continuous monitoring and adjustment. This contrasts with the field-alignment technique, where the underlying dynamics are stable (see \eqref{eq:line_eom_actuator_field_based}), allowing for more intuitive, open-loop control by a human. Consequently, torque/force-based control necessitates digital feedback. This requirement persists even in systems that are passively stable due to intrinsic stiffness, such as catheters, because direct control of torque or force is unintuitive. These inputs correspond to second-order effects (accelerations), rather than directly observable quantities like position or orientation. The applied torques and forces must integrate through the system dynamics first, before producing observable motion. In a closed-loop control architecture, however, the human operator provides high-level commands, such as target orientations, while the feedback controller continuously modulates torques or forces to track these setpoints. This also enables automatic disturbance rejection by the controller, thereby reducing the operator’s workload. By requiring only high-level commands, this approach ultimately becomes even more intuitive for the operator than field-alignment.

\subsubsection{Allocation Strategy}

This paragraph derives the relationship between the controller’s torque outputs and the electrical currents. We present the energy-optimal current allocation for a given motion task, which serves as the primary allocation strategy used throughout the paper for the torque/force-based approach.

As shown in \cite{petruska2015MinimumBounds}, the relationships between torque and magnetic field, $\torque = \m \times \b$, and between force and field gradient, $\force = (\m \cdot \nabla)\b$, can be expressed in matrix form:
\begin{align}
    \< \torque \\ \force \> &= \< \M_\b & \Zero  \\ \Zero & \M_\g \> \< \b \\ \g \> = \M \< \b \\ \g \> \\
    \M_\b &= \skews{\m} = \< 
0 & -\tilde{m}_z & \tilde{m}_y \\
\tilde{m}_z & 0 & -\tilde{m}_x \\
-\tilde{m}_y & \tilde{m}_x & 0 \> \\
    \M_\g &= \< \tilde{m}_x & \tilde{m}_y & \tilde{m}_z & 0 & 0 \\
0 & \tilde{m}_x & 0 & \tilde{m}_y & \tilde{m}_z \\
-\tilde{m}_z & 0 & \tilde{m}_x & -\tilde{m}_z & \tilde{m}_y
 \>.
\end{align}
These expressions highlight that mapping a desired torque/force to a corresponding field/gradient requires knowledge of the magnetic dipole’s current orientation, as the matrix $\M$ depends explicitly on the dipole state $\M = \M(\alpha, \beta)$. Therefore, real-time measurement of the dipole’s orientation is necessary to calculate the required field and gradient for a given torque, as illustrated in \Cref{fig:field_torque_blockdiagram}~b).

Because the actuator of our inverted pendulum is rotationally constrained, any force applied to the magnetic dipole can be converted into an effective torque about the pivot point. We therefore combine both magnetic torques and force-induced torques into a single control quantity $\torque_\c \in \mathbb{R}^3$, defined as $\torque_{\c} = \torque + \tilde{\jacobi} \force$, where $\tilde{\jacobi} = \ell_\emm \skews{\n}$ is the mechanical Jacobian, with $\n = \m/\norm{\m}$ denoting the unit vector along the actuator axis. The vector $\torque_\c$ represents the effective control torques which we define in the body-fixed frame~$\mathcal{B}$ as $\torqueB_\c \coloneqq \< \tau_{\c,x} & \tau_{\c,y} & 0 \>^\T,$ explicitly setting the third component to zero because torques about the dipole axis cannot be generated\footnote{This follows directly from the singular-value decomposition in App.~\ref{app:svd_torque_map}, which reveals that the torque mapping has a one-dimensional nullspace aligned with the dipole axis. Consequently, torques about the dipole axis cannot be generated.}.  Here, $\tau_{\c, x}$ and $\tau_{\c, y}$ are the outputs of the feedback controller. The body-fixed torques are mapped to the inertial frame through $\torque_\c = \R^\T \torqueB_\c$, where $\R \in \mathrm{SO}(3)$ is a rotation matrix. Defining the torque in the body-fixed frame provides both mathematical advantages (see Appendix~\ref{app:svd_torque_map}) and, more importantly, allows operation for any orientation of the dipole, making the control approach general and not limited to specific configurations. While this generality is not critical for our inverted pendulum system, where orientation angles remain small (typically below $10^\circ$) and body-frame and inertial-frame torques are nearly identical, it becomes essential for applications involving large orientation changes or arbitrary dipole configurations.

In general, the Jacobian $\tilde{\jacobi}$ is specific to the mechanical setup and must be adapted accordingly for different systems. Likewise, the actuation matrix $\Act$ depends on the eMNS hardware, while the matrix $\M$ is a general model-based construct applicable to all point-dipoles.

\begin{figure*}[h]
	\centering
	\includegraphics[scale=0.55,   trim = 0.6cm 0.6cm 0.0cm 0.0cm, clip ]{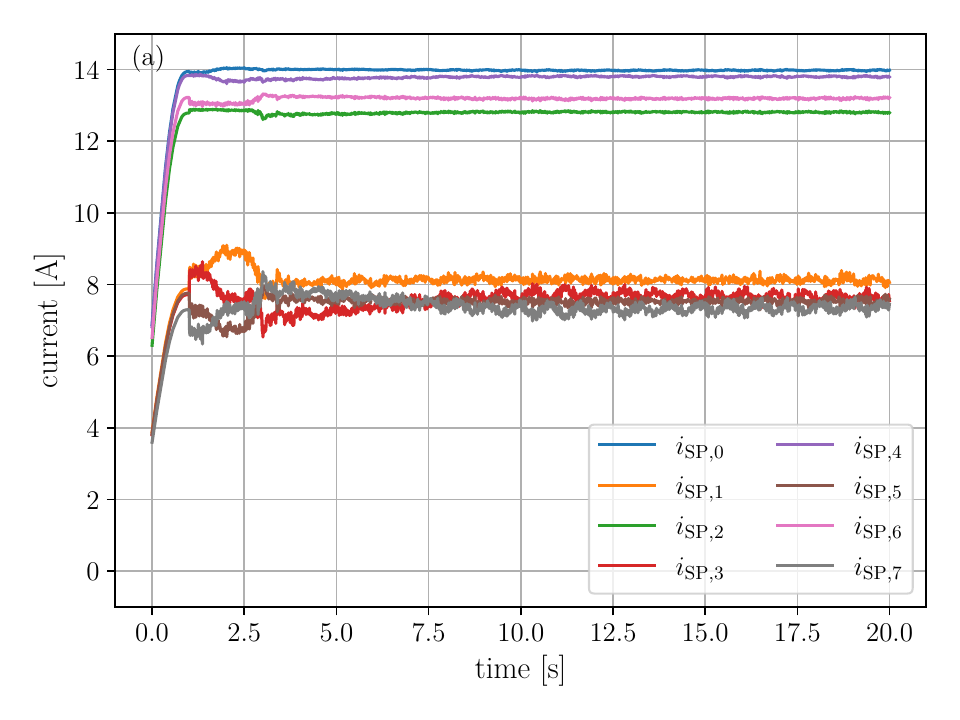} 
    \includegraphics[scale=0.55,   trim = 0.6cm 0.6cm 0.0cm 0.0cm, clip ]{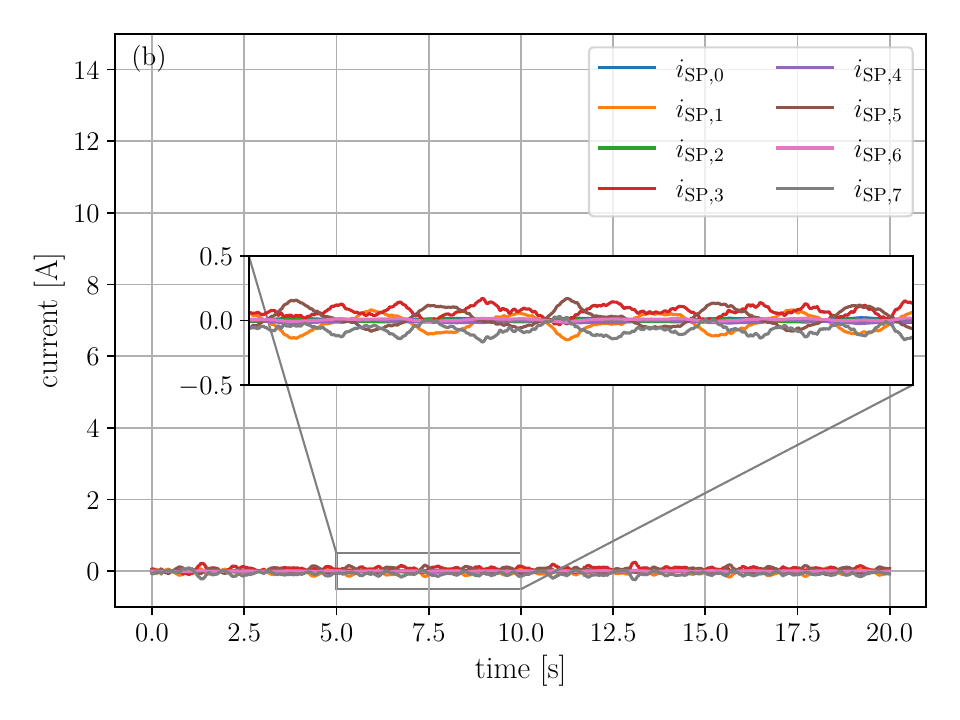} 
	\vspace{-2mm}
    \caption{Electrical currents required to stabilize a single 3D inverted pendulum using the OctoMag eMNS. a): field-alignment control, b): torque/force control with optimal energy allocation. For the same task, the energy-efficient strategy reduces the required currents by up to two orders of magnitude. Although both methods produce comparable forces and torques on the pendulum, field-alignment control wastes significant energy on magnetic field components that do not contribute to actuation. In contrast, optimal energy control ensures the magnetic energy closely matches the mechanical energy required.}
	\label{fig:current_comparison_single_task}
\vspace{\figcapskip}
\end{figure*}

Accordingly, the full mapping from the feedback controller’s output $\torque_\c$ to the coil currents is given by
\begin{align}
    \torque_\c =  \jacobi(\alpha, \beta) \M(\alpha, \beta) \Act(\p)  \current_\SP, \label{eq:torque_to_current_mapping_pend} 
\end{align}
where $\jacobi(\alpha, \beta) \coloneqq \< \I & \tilde{\jacobi} \>$. The required coil currents can be computed using the pseudoinverse, similar to \cite{berkelman2013levitation}:
\begin{align}
    \current_\SP = \< \jacobi(\alpha, \beta) \M(\alpha, \beta) \Act(\p) \>^\dagger \torque_\mathrm{c}. \label{eq:pinv_onestep_pend_alloc_torque}
\end{align}
This solution represents the least-squares optimal current allocation for a given \textit{torque-task}. In contrast to the field-alignment allocation in \eqref{eq:field_based_allocation}, \eqref{eq:pinv_onestep_pend_alloc_torque} does not aim to achieve a specific \textit{magnetic-task}, but directly targets the mechanical task of generating a torque. As a result, the torque/force-based controller enables highly energy-efficient actuation. As illustrated in \Cref{fig:current_comparison_single_task}, the coil currents are up to two orders of magnitude lower than those required by the field-alignment approach, requiring currents in the range of \unit[0.1-0.2]{A}.

The allocation in \eqref{eq:pinv_onestep_pend_alloc_torque} is feasible whenever the motion task, here two effective torque components, lies in the range of the composed map $\jacobi \M \Act$ subject to current constraints. This condition depends on the coil constellation, region of operation, and object-specific Jacobian $\jacobi$. Minimum coil-number bounds for full wrench control are established in \cite{petruska2015MinimumBounds}. Notably, independent control of fields and gradients is not required, as magnetic torques and gradient-induced forces are jointly projected onto the torque task through $\jacobi$. The three-coil Navion illustrates this: although fields and gradients cannot be regulated independently, the torque task remains achievable throughout the demonstrated workspace. A detailed feasibility analysis is provided in Sec.~\ref{sec:workspace_expansion}, where we propose the \emph{Feasibility Margin}, a metric that quantifies the distance to the feasibility boundary imposed by the current constraints.

\subsubsection{Alternative Torque-to-Current Allocations}

In this paragraph, we examine alternative strategies for allocating control torques to electrical currents. Since multiple current configurations can produce the same magnetic torque, the current allocation problem is inherently non-unique. Mathematically, this is due to the non-trivial nullspace of the torque mapping. Our main objective is to emphasize that only the allocation defined in \eqref{eq:pinv_onestep_pend_alloc_torque} provides the energy-optimal solution for a given motion task.

The forward model in \eqref{eq:torque_to_current_mapping_pend} represents a composition of sequential mappings, where currents are first mapped to magnetic fields/gradients through $\Act$, then to torques/forces through $\M$, and finally to torques through $\jacobi$. For the inverse problem, one might intuitively attempt to reverse this sequence step-by-step. However, this approach does not yield the energy-optimal solution. Instead, the minimum-energy solution is obtained only by applying the pseudoinverse to the entire composed matrix $\jacobi \M \Act$ in a single step. To illustrate this principle, consider a simplified case focusing on pure torque-to-current mapping (though the same reasoning applies to the full model in \eqref{eq:torque_to_current_mapping_pend}):
\begin{align}
    \torque = \skews{\m} \Act_\b \current = \M_{\b} \Act_\b \current. \label{eq:tau_i_relation}
\end{align}

Given~\eqref{eq:tau_i_relation}, we can approach the inverse problem in two fundamentally different ways:
\begin{align}
    \current^\aI &= \< \M_{\b} \Act_\b \>^\dagger \torque, 
    &&\text{(one-step allocation)}, \label{eq:onestep_alloc}\\
    \current^\aII &= \Act_\b^\dagger \M_{\b}^\dagger \torque, 
    &&\text{(two-step allocation)}. \label{eq:two_step_alloc}
\end{align}
The one-step allocation computes the entire mapping in a single pseudoinverse operation, while the two-step allocation first determines the minimum-norm field, before mapping it to the currents.

A critical mathematical property underlies this distinction: for two matrices $\bm{C}\in \mathbb{R}^{m \times n}$ and $\bm{D} \in \mathbb{R}^{n \times p}$, the pseudoinverse of their product generally differs from the product of their pseudoinverses, i.e.
$(\bm{C}\bm{D})^\dagger \neq \bm{D}^\dagger \bm{C}^\dagger$
with potentially significant discrepancies for the solution.

According to the Moore–Penrose pseudoinverse theorem~\cite[Thm.~5.5.1]{golub2013matrix}, these two approaches optimize different objectives: the one-step allocation minimizes current magnitude (minimum $\|\current\|$, where $\norm{\cdot}$ denotes the $\ell^2$-norm), while the two-step allocation first minimizes field magnitude (minimum $\|\b\|$) before mapping it to currents. Importantly, these objectives are not equivalent.

A detailed mathematical analysis (see Appendix~\ref{app:allocation_comparison}) reveals the counterintuitive result that despite having a larger field norm $\|\b^\aI\| \geq \|\b^\aII\|$, the one-step allocation requires smaller currents: $\|\current^\aI\| \leq \|\current^\aII\|$. The one-step allocation can exploit the nullspace of $\M_\b$ (parallel-field components) to lower the required coil currents, while the two-step allocation strictly avoids any parallel component to minimize field strength. For the two step allocation, the magnetic field vector remains strictly orthogonal to the dipole moment at all times, as proven by $\m^\T \b = \m^\T \skews{\m}^\dagger \torque = -\abs{\m}^{-2}\m^\T \skews{\m} \torque = 0$. However, only the one-step allocation guarantees minimum current norm for a given torque task, which is why it is used throughout this work for torque/force-based control.

This distinction is not merely theoretical; the choice of torque-to-current allocation method significantly impacts system performance, as illustrated in \Cref{fig:torque_allocation_norm_comparison_app}. It is important to note, however, that even a suboptimal torque-to-current allocation yields currents that are lower than those required by the field-alignment method.

\subsubsection{Expansion of Effective Workspace}
\label{sec:workspace_expansion}

\begin{figure*}[h]
	\centering
	\includegraphics[scale=0.31,   trim = 0.0cm 0.0cm 0.0cm 0.0cm, clip ]{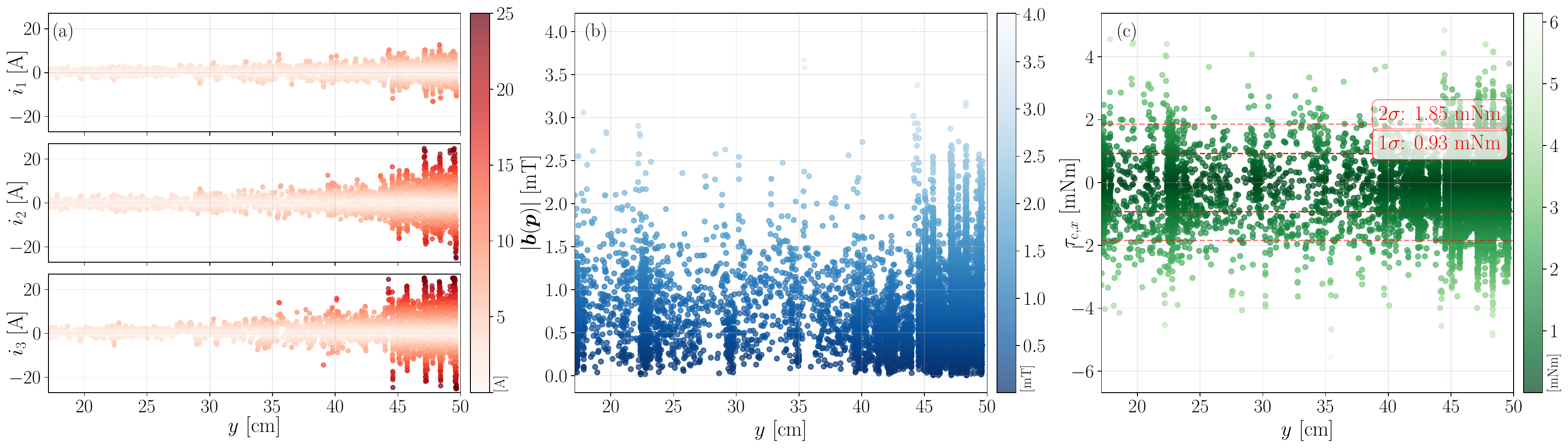} 
    \caption{The figure displays key metrics for stabilizing the inverted pendulum using the Navion eMNS under the torque/force-based control paradigm. Data was collected while the pendulum was gradually moved away from the system, remaining approximately undisturbed in its upright equilibrium. Plot a) shows the three Navion coil currents increasing with distance, as expected. Notably, the magnetic field strengths at the magnet position in b) and the torque amplitudes - representing the feedback controller output - in c) remain roughly constant in peak magnitude, largely unaffected by the distance from the coils. }
	\label{fig:currents_over_positions_navion}
\vspace{\figcapskip}
\end{figure*}

The choice of the torque/force control method using energy efficient allocation is particularly promising to expand the workspace of eMNS for single-agent control task. To illustrate this point, \Cref{fig:navion_inv_pend_moving_around} shows a demonstration of inverted pendulum stabilization using the Navion eMNS while the base is moved manually in space. Stabilization is achieved up to a $y$-distance of \unit[50]{cm} from the coils, under a current constraint of $\norm{\current}_\infty < \bar{i}$, with $\bar{i} = \unit[25]{A}$ (see \Cref{fig:currents_over_positions_navion}). While the current drivers support up to \unit[45]{A}, we conservatively limited the maximum current to \unit[25]{A} for safety reasons - implying that even larger distances could be feasible with the same control strategy.

To quantify the expanded workspace, we adopt an analysis approach inspired by \cite{quentin2023workspace}, but extend it in two key aspects. First, whereas \cite{quentin2023workspace} evaluates feasibility in the task space, we instead perform the analysis in the inverse-mapped current domain. This enables a direct comparison between the two control paradigms, which otherwise operate in different physical units (T and T/m vs. Nm and N); in the current domain, both are expressed uniformly in Amperes. Second, inspired by the DA-distance introduced in \cite{quentin2023workspace}, we propose a novel metric - the Feasibility Margin - which measures how robustly a given position satisfies current constraints. This offers a more informative view on the workspace, moving beyond a simple binary feasible/infeasible classification.

Notably, the required torque to stabilize the inverted pendulum is independent of the position in space. From experimental results, we find that the output of the feedback controllers $\tau_{c,x}$ and $\tau_{c,y}$ remains within the interval $[-\bar{\tau}, \bar{\tau}]$ across the workspace (in upright equilibrium), with $\bar{\tau} = \unit[2.5]{mNm}$, which is the \unit[98]{\%} torque percentile during equilibrium stabilization, as can be seen in \Cref{fig:currents_over_positions_navion}~c).

We define the desired torque task as $\mathcal{D}_\tau = \{ \torque \in \mathbb{R}^3 \mid \norm{\torque}_\infty < \bar{\tau} \}$. Since this task set is convex and the inverse mapping $\current = \<\jacobi \M \Act(\p)\>^\dagger \torque$ is linear (for fixed dipole orientation and position), feasibility can be efficiently evaluated over a discretized grid. This yields the electromagnetic workspace: 
\begin{align*}
    \big\{ \p \in \mathbb{R}^3 |\current = \< \jacobi \M \Act(\p) \>^\dagger \torque,\, \norm{\torque}_\infty < \bar{\tau}\;\text{s.t.}\; \norm{\current}_\infty < \bar{i} \big\}.
\end{align*}
Here, we consider an upright dipole configuration $\m = |\m|\bm{e}_z$, which is representative of the actuator orientation used in pendulum stabilization\footnote{We also analyzed robustness to small angular deviations $\alpha, \beta \in [-5^\circ, 5^\circ]$ using worst-case feasibility checks; as no significant differences were observed, only the upright case is presented.}. 

\begin{figure*}[h]
	\centering
	\includegraphics[scale=0.5723,   trim = 5.9cm 4.5cm 9.445cm 3.5cm, clip ]{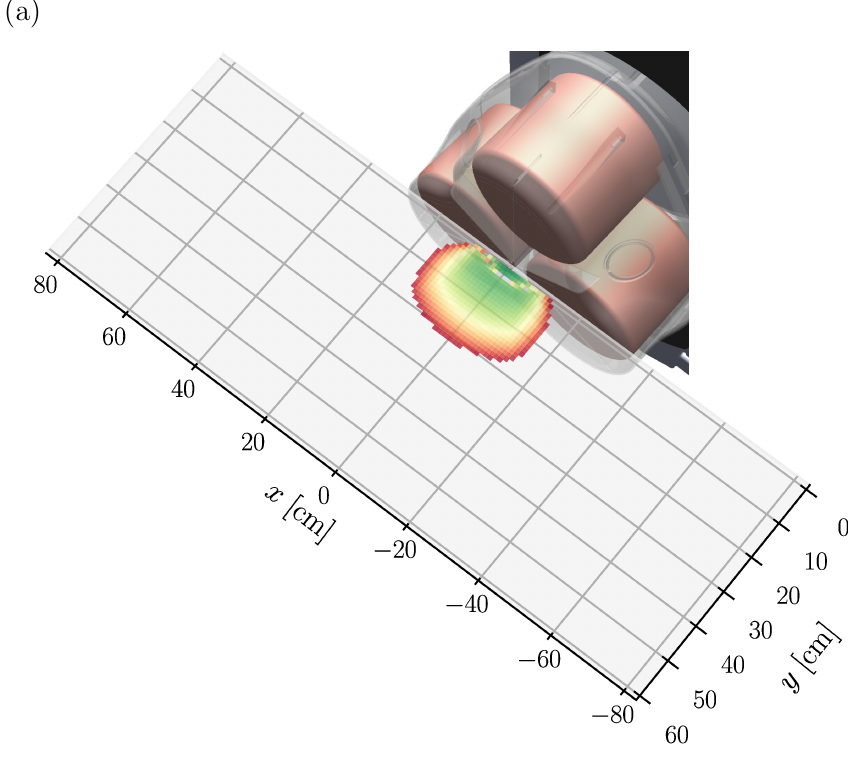}
    \includegraphics[scale=0.5723,   trim = 6.2cm 4.5cm 6.0cm 3.5cm, clip ]{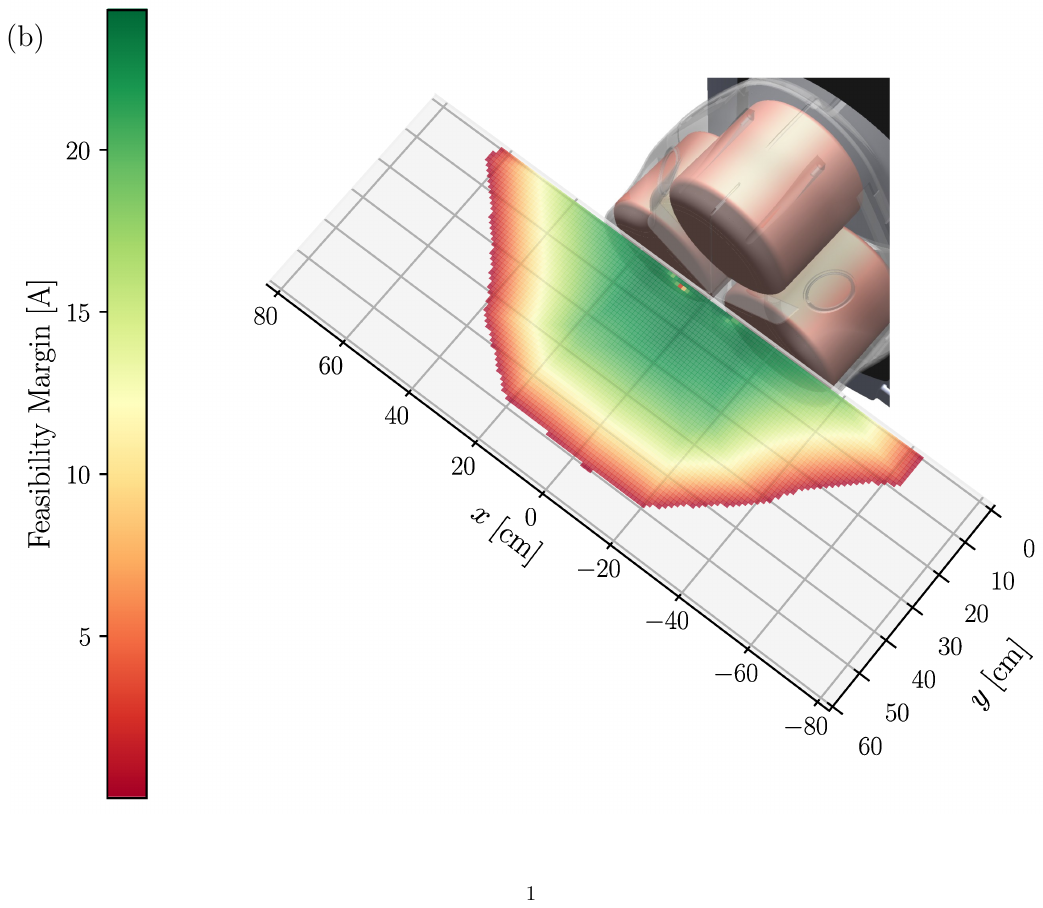} 
	\caption{Electromagnetic workspace for stabilizing the inverted pendulum using field-alignment control a) and energy-efficient torque/force control b). The figure highlights how appropriate structural choices in feedback and allocation dramatically affect performance, and shows that relying on field-alignment for magnetic actuation can be highly limiting. Colors indicate the Feasibility Margin (FM), the available current headroom at each position. A larger FM implies greater robustness and a wider usable range before current constraints become active. Note that the distinct regions of low FM visible near the coil centers in both a) and b) are numerical artifacts resulting from spurious workspace ill-conditioning induced by the Multipole Expansion Model used for calibration~\cite{bernardes2026structured}.}
	\label{fig:workspace_comparison} 
\vspace{\figcapskip}
\end{figure*}

To formalize the Feasibility Margin, FM$(\p)$, we define it as the available current headroom at position $\p$:
\begin{align}
    \text{FM}_{\torque}(\p) = \bar{i} - \max_{\,\norm{\torque}_\infty \leq \bar{\tau}} \big\|\,\< \jacobi \M \Act(\p) \>^\dagger \torque\,\big\|_\infty, \label{eq:feasibility_margin_torque}
\end{align}
which has units of Amperes and consequently has the advantage that it is comparable across the two different paradigms. We use $\text{FM}_{\torque}(\p)$ to compare the field-alignment and torque/force-based workspaces. For field-alignment, the Feasibility Margin is defined with respect to the task set $\mathcal{D}_b = \{ \b \in \mathbb{R}^3 \mid \b = |\b| \, \bm{e}_z \}$, i.e.:
\begin{align}
    \text{FM}_\b(\p) = \bar{i} - \max_{\,\b \in \mathcal{D}_b} \big\|\,\Act_{\b}(\p)^{\dagger} \, \b\,\big\|_\infty. \label{eq:feasibility_margin_field}
\end{align}
To ensure a fair comparison, both task sets must demand equivalent baseline control authority. In the field-alignment paradigm, a magnetic field provides a restoring stiffness of $|\m||\b|$ against angular deviations of the dipole (see linearized dynamics in \eqref{eq:line_eom_actuator_P_ctrl}). Conversely, the active restoring stiffness of the torque controller is governed by the proportional gain on the actuator angle $\alpha$ (i.e. the first element of the LQR gain matrix $\K_\alpha$ in \eqref{eq:control_policy}). By setting $|\b| = k_{\alpha, p} / |\m|$, we match the open-loop magnetic stiffness to the active closed-loop digital stiffness. This mathematical equivalence ensures that the workspace analysis evaluates both methods under functionally identical performance criteria. The resulting equivalent field strength $|\b|$ evaluates to \unit[12.9]{mT}. The corresponding workspace for the field-alignment paradigm is given by
\begin{align}
    \big\{ \p \in \mathbb{R}^3 \mid \current = \Act_{\b}^\dagger(\p) \, \b,\; \b = |\b| \, \bm{e}_z\;\text{s.t.}\; \|\current\|_\infty < \bar{i} \big\}.
\end{align}
\Cref{fig:workspace_comparison} illustrates both workspaces, demonstrating the significantly larger feasible region enabled by the energy-efficient control approach.

\subsubsection{Influence of Sampling Rate and Delays}
\begin{figure*}[htbp]
    \centering
        \includegraphics[width=0.46\textwidth]{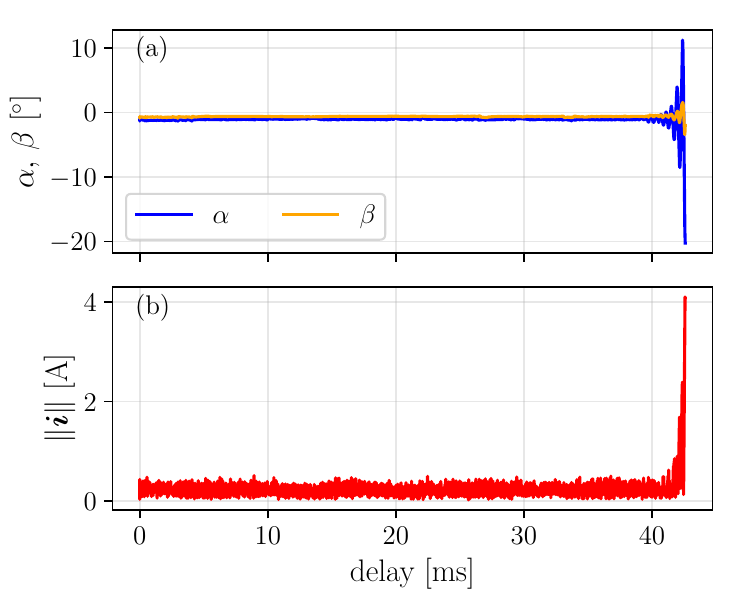}
        \includegraphics[width=0.46\textwidth]{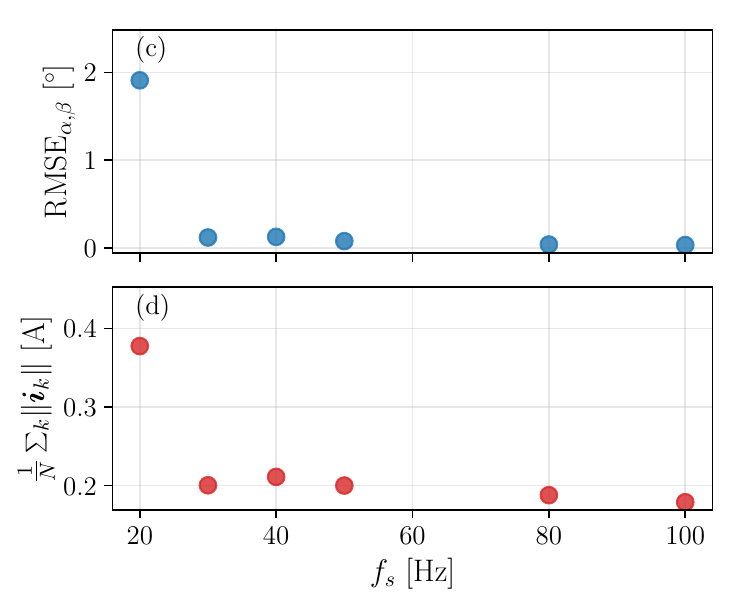}
    \caption{Influence of artificially injected communication delays (a, b) and different sampling rates (c, d) on the performance of the actuator (no pendulum attached) using the Navion system. Throughout these experiments, the actuator is stabilized in the upright equilibrium (no pendulum attached). For (a, b), the delay is continuously increased over a 1-minute time period. For (c, d), the root-mean-square error (RMSE) is evaluated over 10-second intervals across different sampling frequencies. The results demonstrate that the required coil currents remain unaffected across a wide range of delays and reduced sampling frequencies. Accumulated phase lag eventually destabilizes the (unstable) system at $\approx$\unit[40]{ms} of delay, and marginal stability is reached at a sampling rate of $\approx$\unit[20]{Hz}. }
    \label{fig:frequency_and_delay_influence}
\vspace{\figcapskip}
\end{figure*}

In clinical practice, state-of-the-art localization modalities often exhibit limited sampling frequencies and inherent communication delays; hence, we demonstrate that our methodology is robust against such constraints. The exact sampling rate and maximum allowable delay fundamentally depend on the specific task and its underlying system dynamics. For the delay and sampling rate analysis, we evaluate the actuator without the pendulum attached. Its open-loop unstable dynamics (see \eqref{eq:actuator_eom_no_pend_torque_force}) intrinsically demand a sufficiently high sampling rate and strictly bound the permissible delay to maintain closed-loop stability. Nonetheless, as demonstrated in \Cref{fig:frequency_and_delay_influence}, as long as the system operates within its stability bounds, the energy-optimal current allocation remains unaffected by reduced sampling rates and injected delays. Consequently, for passively stable systems like clinical catheters, it is expected that even lower frequencies and higher delays can be handled without affecting performance and energy efficiency.

\label{sec:Influence of Sampling Rate and Delays}

\subsection{Discussion}

Comparing the electrical currents required to stabilize the inverted pendulum using field-alignment control and torque/force-based control, we can notice that the field-alignment control with digital feedback can stabilize the system, but at the cost of substantial energy consumption. Indeed, while field-alignment requires currents in the range of \unit[8–14]{A}, the torque/force-based approach with optimal energy allocation achieves the same stabilization task with currents as low as \unit[0.1–0.2]{A}. To understand the origin of this disparity, it is helpful to consider the torque equation in a simplified 2D case:
\begin{align}
\tau = \mb \sin(u_\alpha - \alpha), \label{eq:2D_torque_equation}
\end{align}
where $\alpha$ denotes the orientation of the magnetic dipole and $u_\alpha$ the orientation of the magnetic field vector. As the term \textit{field-alignment} suggests, the dipole tends to align with the applied magnetic field, resulting in a small angular difference $u_\alpha - \alpha \approx 0$. Consequently, $\sin(u_\alpha - \alpha) \approx 0$, and a significant field magnitude $|\b|$ (or magnetic volume $|\m|$) is required to generate a significant torque $\tau$.

In contrast, the energy-efficient control technique does not rely on a small-angle assumption between the field and the dipole. Instead, the field orientation emerges naturally from the allocation process as a mathematical consequence of minimizing electrical energy. The allocation consistently produces field vectors that remain approximately orthogonal to the dipole, as demonstrated in \Cref{fig:torque_allocation_norm_comparison_app}. With this orthogonal configuration, the sinusoidal term in \eqref{eq:2D_torque_equation} evaluates to $\approx1$, allowing the same torque $\tau$ to be achieved with a substantially smaller field magnitude or magnetic volume. 

This fundamental difference in field orientation directly explains the order-of-magnitude improvement in energy efficiency. In field-alignment control, a significant portion of the magnetic energy is concentrated in the component $\b \parallel \m$, which lies in the nullspace of the torque map and therefore cannot produce actuation (see App.~\ref{app:svd_torque_map} for a singular-value-decomposition of the torque map). The energy invested in this parallel component is effectively wasted. In contrast, energy-efficient control strategically minimizes the parallel field component, permitting its presence only when it results in lower overall electrical energy consumption, as explained in the one-step versus two-step allocation analysis above and detailed in App.~\ref{app:allocation_comparison}. This approach ensures that the magnetic energy is transferred into mechanical energy with optimal efficiency.

This improved efficiency has several practical implications. For the same motion task, one can:

\begin{itemize}
    \item Operate a smaller eMNS (potentially avoiding the need for active cooling),
    \item Use magnetic agents with significantly lower magnetic volume,
    \item Expand the effective workspace, since electrical current constraints are reached at greater distances from the coils.
\end{itemize}

This last point was illustrated using the Navion eMNS. It is worth noting that even at \unit[25]{mT}, the field-alignment strategy does not guarantee zero steady-state error in the absence of digital feedback. Due to the coil configuration of the Navion system, with coils positioned laterally, a strong gradient field exists that tends to pull the magnetic object toward the coils. This arises from the $1/r^3$ decay of magnetic fields with distance. While OctoMag's eight-coil configuration allows for the generation of approximately uniform fields (by setting the field gradient $\g = \Zero$), the Navion eMNS cannot achieve this.

This highlights another advantage of the torque/force-based strategy, which is that it inherently accounts for field gradients by explicitly incorporating magnetic forces into the control objective. While gradient compensation is theoretically possible with field-alignment control, it is more involved and sensitive to the validity of small-angle assumptions. As shown in App.~\ref{sec:appendix_gradient_incorp}, the method breaks down when gradients are strong enough to violate these assumptions.

\section{Multi-Agent Control}
\label{sec:Multi-Task_Control}

In this section, we extend the proposed approach to the control of multiple agents within the same workspace, by exploiting magnetic-field nonlinearities and coil redundancy to independently actuate multiple identical agents. All experiments in this section are conducted with the OctoMag system, whose eight coils provide the redundancy required for independent actuation of two magnetic agents.

Controlling the multi-agent configuration is more challenging due to inherent coupling effects. In particular, two primary sources of coupling arise: (1) magnetic interactions between the permanent magnets of the two pendulum systems, and (2) cross-talk from the shared electromagnetic coils, which simultaneously influence both magnetic dipoles. Since magnetic field strength decays cubically with distance, each coil predominantly affects the closer agent. This spatial decay is the key property that enables feasible multi-agent control within a shared magnetic workspace.

In the multi-agent control setting, we apply the field-alignment strategy only to the 2$\times$2D configuration for simplicity. While this approach is, in principle, extendable to the 2$\times$3D case, we limit its application to the 2$\times$2D setup due to practical considerations - most notably, thermal constraints on the coils and driver electronics\footnote{We verified that two 3D actuators can be feedforward stabilized within the OctoMag workspace.}. For the more demanding 2$\times$3D case, we instead adopt the torque/force-based approach, which scales more naturally and remains efficient in terms of current usage.

\subsection{Field-Alignment Control}
\label{sec:multi_task:field_align}
\begin{figure*}
	\centering
	\includegraphics[scale=0.55,   trim = 0.0cm 0.2cm 0.0cm 0.0cm, clip ]{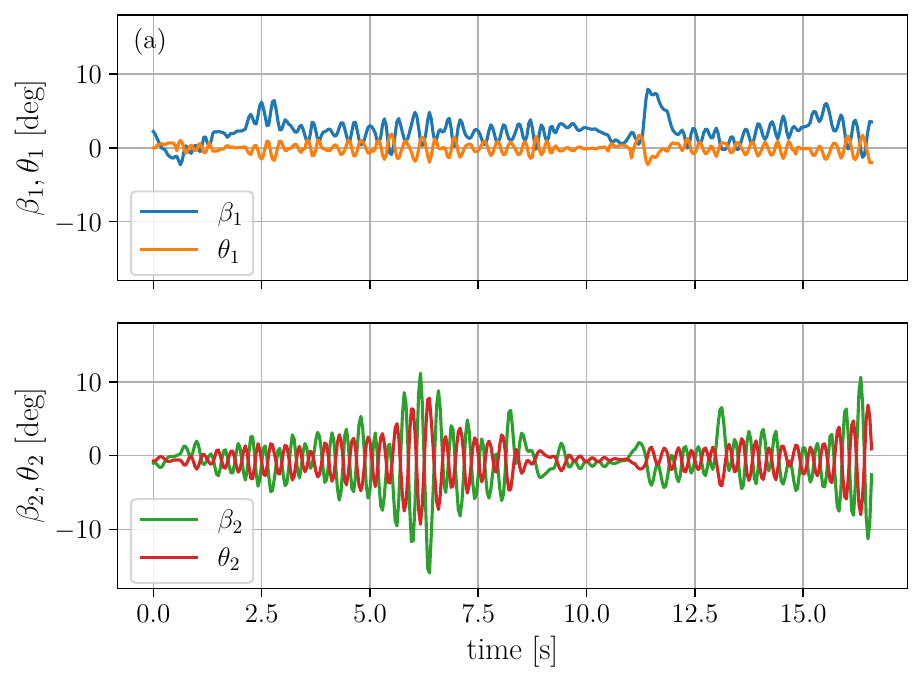} 
	\includegraphics[scale=0.55,   trim = 0.0cm 0.2cm 0.0cm 0.0cm, clip ]{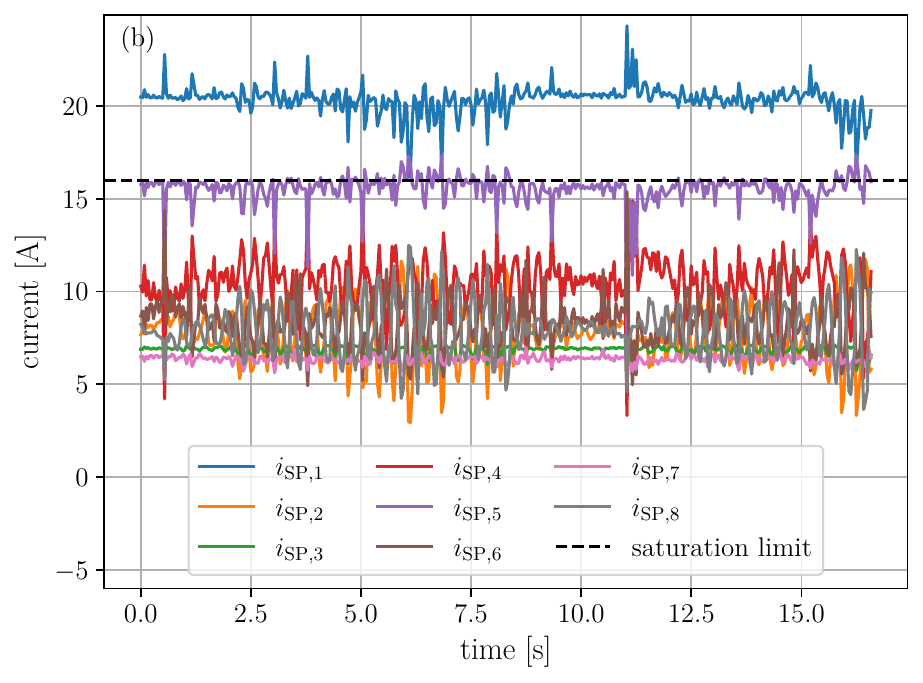} 
	\vspace{-2mm}
	\caption{Stabilization of the 2$\times$2D inverted pendulum setup using field-alignment control.
	a) Time evolution of both pendulum angles, showing stable upright balancing. Manual perturbations introduce the visible oscillations. No integral control is used in this experiment.
	b) Corresponding actuator currents computed by the field-alignment controller operating at a magnetic field strength of \unit[65]{mT} at each of the two positions in the workspace. Although one coil permanently exceeds the saturation limit of \unit[16]{A} (i.e. only \unit[16]{A} are applied), the system remains stable due to the inherent actuator redundancy in the 2$\times$2D configuration and the availability of feedback which allows to compensate for unmodeled disturbances.}
	\label{fig:2x2D_pendulum_field_analysis}
\vspace{\figcapskip}
\end{figure*}

Let $\b_{\SP, 1}$ and $\b_{\SP, 2}$ be two desired magnetic field setpoints to be generated at positions~$\p_1$ and~$\p_2$ respectively, and determined by feedback controllers $C_1$ and $C_2$. The corresponding currents are calculated as
\begin{align}
   \current_\SP = \< \Act_{\b}(\p_1) \\ \Act_{\b}(\p_2) \>^\dagger \< \b_{\SP, 1} \\ \b_{\SP, 2} \> \label{eq:multi_task_field_alignment_allocation}
\end{align}
where $\Act_{\b}(\p_i) \in \mathbb{R}^{3 \times 8}$ is the field-actuation matrix at position $\p_i$. \Cref{fig:2x2D_pendulum_field_analysis} presents experimental results for the stabilization of the 2$\times$2D inverted pendulum setup. We were able to achieve stabilization using a field-magnitude $\abs{\b_{\SP, 1}} = \abs{\b_{\SP, 2}} = \unit[65]{mT}$ at each of the two positions in the workspace. The corresponding coil currents are shown in \Cref{fig:2x2D_pendulum_field_analysis}~b). A demonstration of the successful 2x2D stabilization is included in the supplementary video.

\subsection{Torque/Force-Based Control}

Let $\torque_{\mathrm{c},1}$ and $\torque_{\mathrm{c},2}$ be two control torques to be generated on magnets attached to the actuators located at positions~$\p_1$ and~$\p_2$ respectively. These are given by two feedback controllers $C_1$ and $C_2$, so that the desired coil currents computed through the minimum-energy allocation become
\begin{align}
    \current_\SP = \< \jacobi(\alpha_1, \beta_1) \, \M(\alpha_1, \beta_1) \, \Act(\p_1) \\
    \jacobi(\alpha_2, \beta_2) \, \M(\alpha_2, \beta_2) \, \Act(\p_2) \>^\dagger \< \torque_{\c, 1} \\ \torque_{\c, 2} \>. \label{eq:multi_task_torque_allocation}
\end{align}
Note that the computational complexity of this allocation scales solely with the number of commanded torques rather than the system's internal degrees of freedom, preserving real-time feasibility even for future extensions to continuum systems.

\begin{figure*}
	\centering
	\includegraphics[scale=0.305,   trim = 0.0cm 0.0cm 0.0cm 0.0cm, clip ]{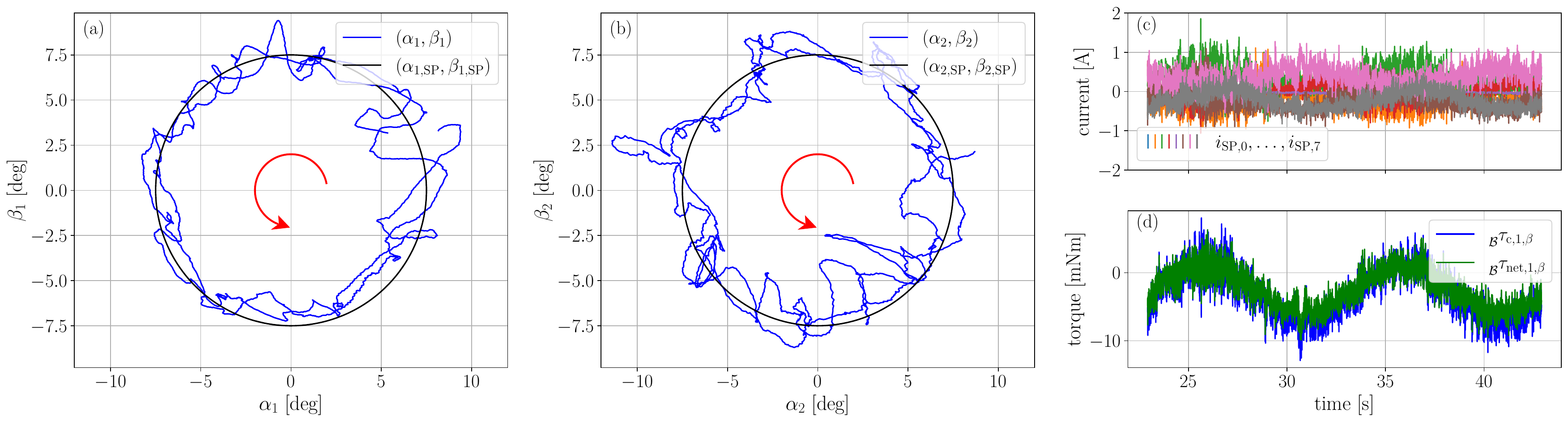} 
	\caption{Trajectory tracking results for simultaneous stabilization of the 2$\times$3D inverted pendulums  within a shared magnetic workspace. These results are obtained using the torque/force-based paradigm. a) and b): Experimental tracking performance at two different positions (position 1 and position 2), where both actuators rotate in-phase at equal frequency in the counterclockwise direction. Similar results are obtained when the actuators rotate out-of-phase (i.e., one clockwise, one counterclockwise). We can also achieve stable tracking even when the two pendulum systems are configured with opposing magnetic polarities (north-pole up vs. south-pole up). These results are achieved using a fully decoupled LQRI architecture, where each degree of freedom is controlled independently. Future work may improve tracking further by incorporating coupling effects-both magnetic interaction between the pendulums and cross-coupling through the shared electromagnets. c): The corresponding coil currents for the experiment. As expected, current demands are higher than in the single-agent setting, but remain significantly lower than field-alignment control. d): Comparison between the nominal control torque and the actual net torque for object 1 along the $\beta$ orientation. To account for spatial field variations and gradient-induced forces, the net torque is evaluated by discretizing the magnetic volume and aggregating the individual torque contributions, as described in~\cite{zughaibi2024balancing}. }
	\label{fig:lissajous_2x3D_pend_currents}
\vspace{\figcapskip}
\end{figure*}

\begin{figure*}
	\centering
	\includegraphics[scale=0.305,   trim = 0.0cm 0.0cm 0.0cm 0.0cm, clip ]{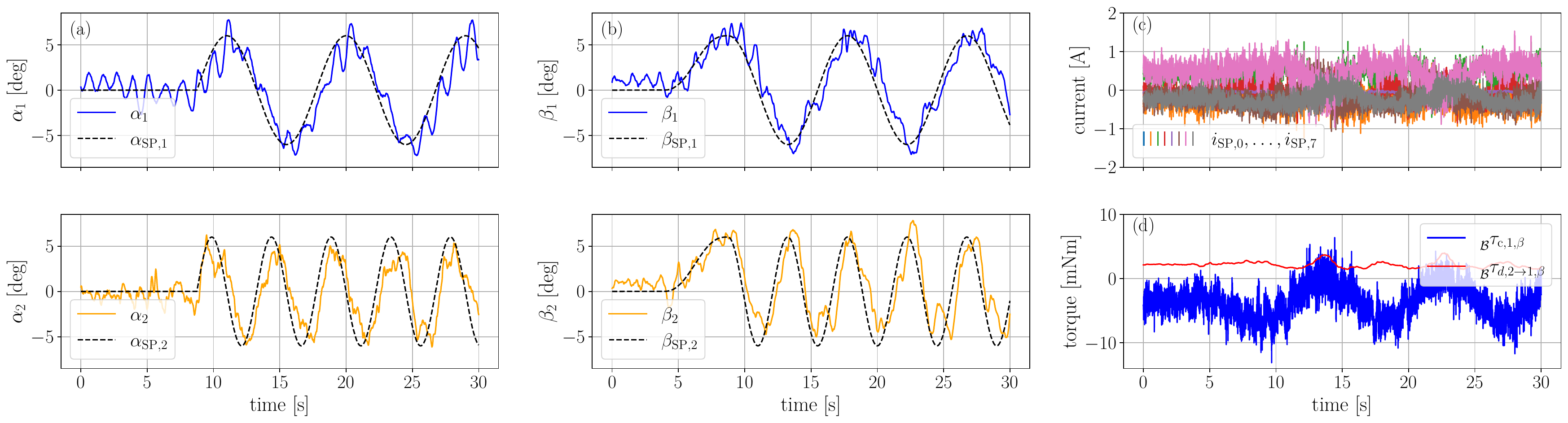} 
	\caption{Trajectory tracking results for simultaneous stabilization of two 3D inverted pendulums within a shared magnetic workspace using the torque/force-based control paradigm. In contrast to the synchronous case shown in \Cref{fig:lissajous_2x3D_pend_currents}, this experiment demonstrates the system’s ability to track two circular trajectories evolving at distinct frequencies. While stability is preserved, tracking accuracy slightly degrades likely due to increased control complexity and interaction effects between agents operating on different time scales. These limitations could be mitigated with more advanced controllers, but we intentionally use simple algorithms to highlight the role of structural design choices. The corresponding coil currents in c) remain within similar bounds as in the equal-frequency case. d) Comparison of the applied control torque and the estimated inter-agent magnetic disturbance torque, demonstrating that the parasitic coupling effects remain bounded and are actively rejected by the decoupled controllers. } 
	\label{fig:lissajous_2x3D_pend_currents_async}
\vspace{\figcapskip}
\end{figure*}

\Cref{fig:lissajous_2x3D_pend_currents} illustrates the application of this allocation to the simultaneous trajectory tracking of the $2\times$3D pendulum system within a common magnetic workspace. In this case, both actuator setpoints are commanded in-phase at the same frequency. \Cref{fig:lissajous_2x3D_pend_currents_async} presents similar experiments for asynchronous tracking, where each pendulum follows a circular trajectory at a different frequency - one rotating twice as fast as the other. Although stability is maintained, tracking performance degrades slightly, likely due to the increased control complexity. In both cases, the coil currents remain within comparable bounds of approximately $\pm$\unit[1]{A}.

The supplementary video showcases the 2$\times$3D pendulum experiments in action. Notably, the system remains stable even when the pendulums are configured with opposing magnetic polarizations (i.e., one north-pole up, one south-pole up). These results are achieved using a fully decoupled LQRI control architecture, with each degree of freedom regulated independently. Future enhancements will incorporate cross-coupling effects - both magnetic interactions and coil-induced coupling - into the feedback controller design to improve tracking performance.

To successfully stabilize the 2$\times$3D pendulums, several practical considerations must be addressed - such as the sequential release of the pendulums and offset compensation. These are elaborated in App.~\ref{app:practical_considerations}, which also includes a discussion of the primary sources of steady-state errors observed during operation.

\subsection{Expansion of Effective Workspace}

\begin{figure*}
	\centering
	\includegraphics[scale=0.24,   trim = 0.0cm 0.0cm 0.0cm 0.0cm, clip ]{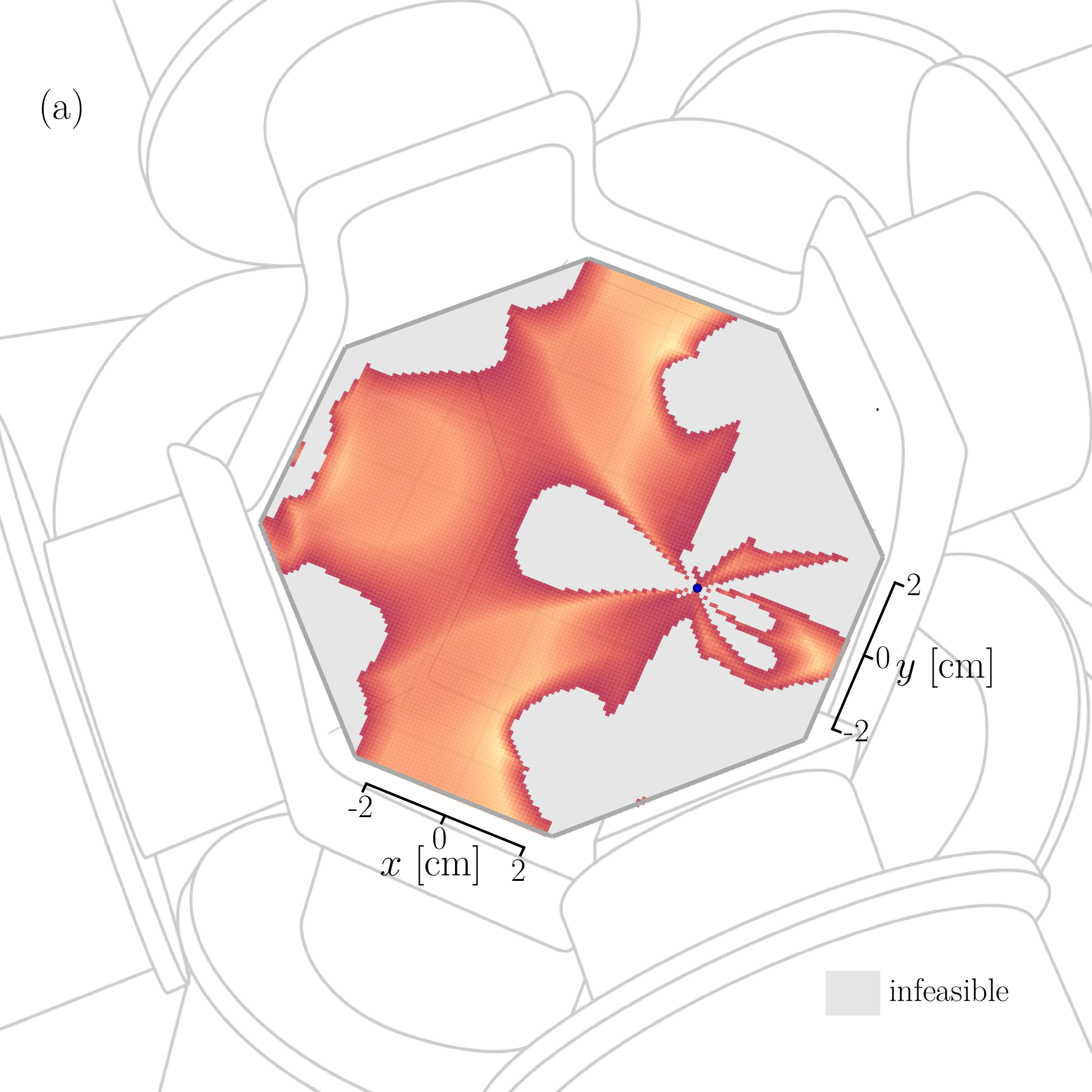} 
	\includegraphics[scale=0.24,   trim = 4.9cm 0.0cm 0.0cm 3cm, clip ]{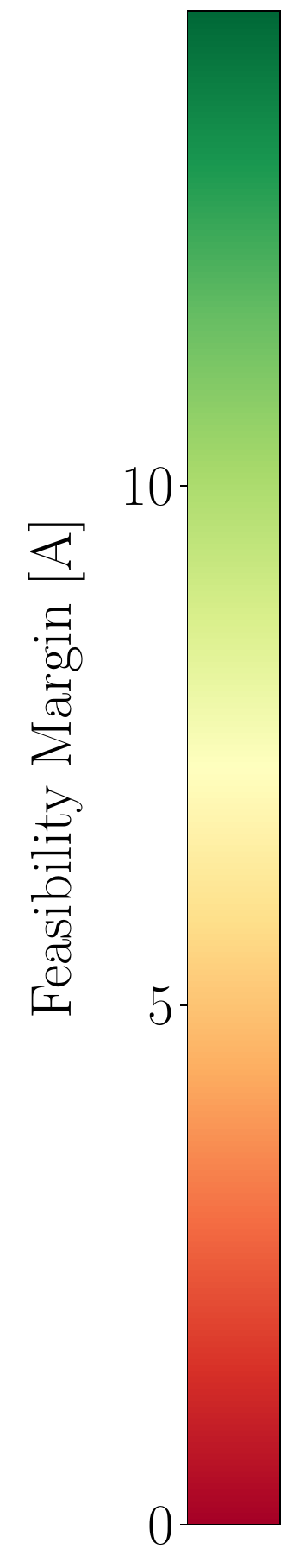} 
	\includegraphics[scale=0.24,   trim = 0.0cm 0.0cm 0.0cm 0.0cm, clip ]{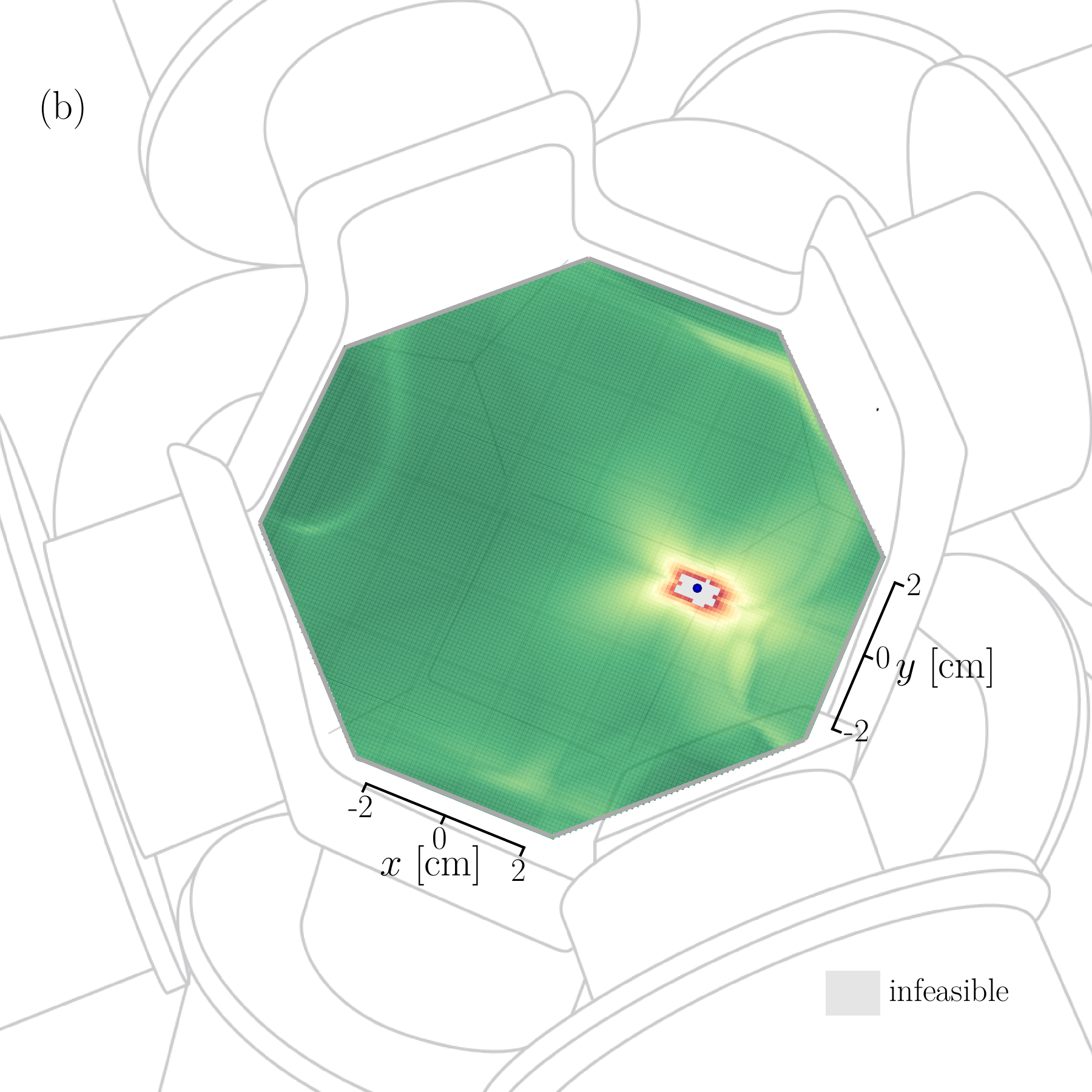} 
	\vspace{-2mm}
    \caption{The figure compares the achievable workspace for controlling two agents in the OctoMag eMNS using (a) field alignment and (b) torque/force-based control with optimal energy allocation. The position $\p_2$ (blue dot) is fixed while $\p_1$ is varied. A configuration is labeled feasible if all coil currents remain within system limits. Grey regions indicate infeasibility, while color intensity encodes the margin to current saturation. Method (b) achieves a substantially larger feasible workspace. Limitations: The analysis assumes point dipoles, neglecting spatial field variations across each magnet's volume. Inter-magnet interactions are also ignored, potentially overestimating feasibility in close-proximity configurations. Additionally, the multi-pole expansion used to derive $\Act(\p)$ introduces inaccuracies, particularly near the coils. }
	\label{fig:workspace_comparison_multi_task}
    \vspace{-2mm}
\vspace{\figcapskip}
\end{figure*}

As in the single-agent case, we analyze the achievable workspace for controlling two agents in the OctoMag system using both field-alignment and torque/force-based control. However, unlike the single-agent case, the allocation strategies in \eqref{eq:multi_task_field_alignment_allocation} and \eqref{eq:multi_task_torque_allocation} depend on both agent positions $\p_1$ and $\p_2$, making the workspace analysis inherently higher-dimensional. To visualize the results, we fix the position of the second agent at $\p_2 = \bar{\p}_2$. Given a fixed position $\bar{\p}_2$, the feasible workspace for the field-alignment method is defined as
\begin{align*}
    \bigg\{ \p_1 \in \mathbb{R}^3\, \bigg| \current = \< \Act_{\b}(\p_1) \\ \Act_{\b}(\bar{\p}_2) \>^\dagger \< \b_{1} \\ \b_{2} \>,\; \b_i = |\b| \, \bm{e}_z,\,\|\current\|_\infty < \bar{i} \bigg\}.
\end{align*}
Similarly, the workspace for the torque/force-based method is given by
\begin{align*}
    \bigg\{ \p_1 \in \mathbb{R}^3\, \bigg| \current = \< \jacobi \M \Act(\p_1) \\
    \jacobi \M \Act(\bar{\p}_2) \>^\dagger \< \torque_{1} \\ \torque_{2} \>,\, \norm{\torque_i}_\infty < \bar{\tau},\,\norm{\current}_\infty < \bar{i} \bigg\}.
\end{align*}
\Cref{fig:workspace_comparison_multi_task} compares the resulting feasible workspaces for both methods, labeling a configuration as feasible if all coil currents stay within hardware limits. To further characterize robustness, we compute the Feasibility Margin, which quantifies the sensitivity of each configuration to violating current constraints. The Feasibility Margin is formally defined in \eqref{eq:feasibility_margin_torque} and \eqref{eq:feasibility_margin_field} for the single-task setting. We use the same definitions here, with the sole modification that the margins are evaluated using the stacked matrices appearing in the pseudoinverse mappings; explicit formulas are omitted for brevity. The results show that the torque/force-based method consistently achieves a larger workspace compared to field alignment.

We note that the analysis is somewhat idealized. Although configurations with $\p_1$ close to $\bar{\p}_2$ are generally infeasible due to excessive current demands, the model may still predict feasibility in some near-contact cases that are not physically realizable. This discrepancy arises from the assumption of ideal point dipoles, i.e., magnetic moments concentrated at a single point with no spatial extent, thereby neglecting field variation across the magnet volumes. As detailed in~\cite{zughaibi2024balancing}, an accurate assessment of the net torque can be obtained by integrating the spatially varying magnetic fields and gradient-induced forces over the magnetic volume. Despite its simplicity, the point-dipole assumption remains a practically justified approximation for general operation within the workspace. As demonstrated in \Cref{fig:lissajous_2x3D_pend_currents}d), the spatial modeling discrepancy between the nominal control torque and the actual net torque evaluated over the magnetic volume yields a relative root-mean-square error (RMSE) of approximately 17\%. This unmodeled discrepancy acts as a bounded disturbance that is robustly rejected by the feedback controller without compromising stability.

Additionally, for the workspace analysis, inter-magnet interactions are not taken into account, potentially leading to an overestimation of feasibility in close-proximity configurations. To quantify these coupling effects during dynamic operation, we estimate the parasitic inter-agent disturbance forces and torques using the point-dipole approximation. The full mathematical formulation for this disturbance estimation is provided in App. \ref{app:magnetic_disturbance}. As illustrated in \Cref{fig:lissajous_2x3D_pend_currents_async}d), this estimated inter-agent disturbance torque $\torqueB$ remains bounded. In steady-state, the integral action within the feedback controller naturally accumulates this approximately constant disturbance which is compensated by applying a steady-state, opposite-sign control torque (refer to App. \ref{app:practical_considerations}). It should be noted that bringing the agents into closer proximity causes the commanded currents to increase rapidly; this is driven by both the $\mathcal{O}(1/\|\mathbf{r}_{ij}\|^3)$ and $\mathcal{O}(1/\|\mathbf{r}_{ij}\|^4)$ scaling of inter-magnet disturbances in Eqs. \eqref{eq:dist_torque}--\eqref{eq:dist_force}, and the progressive ill-conditioning of the allocation matrix in \eqref{eq:multi_task_torque_allocation}. Ultimately, the minimum operable distance is fundamentally limited by the physical hardware's current capacity. A quasi-static analysis accounting for both of these compounding effects reveals that, given our hardware limit of \unit[4]{A}, a theoretical minimum separation distance of approximately \unit[40]{mm} is achievable. It is worth noting that the \unit[4]{A} hardware limit is chosen conservatively for safety reasons, and relaxing this constraint would facilitate closer operation. Moreover, the calculated \unit[40]{mm} distance reflects the threshold at which a single current constraint becomes active. An advanced allocation method that explicitly incorporates these constraints could exploit coil redundancy to further reduce this separation, presenting an interesting avenue for future research.

While the study considered only two agents, the proposed method generalizes to more. The maximum number of independently controllable agents depends on the coil count, their spatial arrangement, and task-specific constraints. For a single agent, a comprehensive analysis of the theoretical minimum hardware bounds is provided in~\cite{petruska2015MinimumBounds}. While that study notes that full spatial wrench control (for arbitrary orientations) strictly requires at least seven coils per unconstrained agent to avoid static singularities in the allocation matrix, exploiting the restricted angular range of our inverted pendulum systems and dynamic feedback control enables the independent actuation of two agents using only an eight-coil system, and potentially fewer.

\section{Conclusion}
\label{sec:Discussion}

In this work, we demonstrate that appropriate feedback control enables powerful capabilities in electromagnetic navigation, including an expanded workspace in single- and multi-agent settings. We systematically compare two core control paradigms in electromagnetic navigation - field-alignment and torque/force-based control - and demonstrate, through single- and multi-agent experiments, that the latter can reduce actuation energy by up to two orders of magnitude without sacrificing performance. Within the family of torque/force-based methods, we provide a comprehensive analysis of allocating required torques and forces to coil currents and demonstrate the counterintuitive result that the orthogonal field solution does not, in general, minimize energy. Our key contributions include the first independent stabilization of two identical 3D inverted pendulums within a shared magnetic workspace and the stabilization of a pendulum at large stand-off distances from a clinical-grade eMNS, demonstrating a substantial expansion of the effective workspace.

These results stem from the following architectural decisions:
\begin{itemize}
    \item Using torque- and force-level inputs that directly encode the motion task, rather than the intuitive but inefficient field-alignment approach.
    \item Energy-optimal current allocation that keeps the magnetic-field energy at the level required by the mechanical task.
    \item Real-time estimation ($>$\unit[20]{Hz}) of each agent’s orientation and position, as allocating torques and forces to coil currents depends on the dipole orientation and the position-dependent actuation matrix.
    \item Feedback control to stabilize the dynamics and rapidly reject disturbances and effects which are difficult to model (such as errors in the actuation matrix $\Act$).
    \item A high-bandwidth electromagnetic system that provides sufficient torque authority across the relevant frequency range.
\end{itemize}
Our results demonstrate that stronger magnetic fields do not inherently enhance performance. Although large fields benefit field-alignment control by acting as high proportional gains, we show that much of the magnetic energy in this approach does not contribute to motion. Instead, we reformulate the task using torque- and force-based objectives, allocating coil currents to minimize electrical power, which ensures that actuation is directed toward the torque-generating components of the magnetic field. This strategy aligns magnetic energy with the mechanical energy required for the motion task, causing current constraints to become active only at greater stand-off distances. As a result, the same mechanical task remains feasible farther from the source, effectively enlarging the attainable workspace.

Our results highlight a broader design lesson: the principle of parsimony. Also known as Occam’s razor and widely used in the field of machine learning, it favors the simplest effective approach. When a foundational architectural choice is suboptimal (e.g., adopting field-alignment for dynamic stabilization), even the most sophisticated algorithm cannot overcome the resulting structural limitations; a system is only as strong as its weakest link. It is important to note that all of the architectural decisions above are equally essential. In that light, simple, computationally light linear dynamical models, paired with high-frequency feedback, are often sufficient and remarkably powerful. More elaborate control and modeling architectures are justified only if they preserve these principles - e.g., by meeting strict real-time constraints - and deliver clear, measurable benefits.

From a broader perspective, the control principles established here are scale-agnostic, which has important implications for related fields such as microrobotics. While our experiments focus on macro-scale pendulums within a clinical-grade eMNS, the same challenges of power constraints and workspace limitations apply at the micro-scale. For example, applying our energy-optimal torque paradigm to drive permanent-magnet helical microrobots \cite{reviwer2024} could significantly reduce coil currents and expand operational workspaces. Implementing this dynamic current allocation fundamentally depends on the availability of the instantaneous dipole orientation. This reveals a core challenge in the field: while a large amount of microrobotics literature relies on optical feedback within constrained, petri-dish environments, clinical translation requires operation within deep human tissue where line-of-sight is lost. Therefore, advancing real-time, non-visual state estimation for microscale agents is both a fundamental problem and a promising future direction. Overcoming this estimation bottleneck will not only unlock efficient control paradigms, but also bridge the gap between benchtop demonstrations and human-scale clinical applications.

In this work, we deliberately employed simple, standard control schemes such as LQR and integral action to highlight the impact of structural decisions at the systems level. Future work will explore more advanced control strategies that account for coupling effects in multi-agent scenarios.

{\appendices
\section{Incorporating Gradients in Field-Alignment Control}
\label{sec:appendix_gradient_incorp}
Here we illustrate how to include field gradients so that these explicitly enter the equations of motion. This shows that incorporating gradient-based effects directly can be more involved compared to the torque/force-based method (see Sec.~\ref{sec:Torque/Force-based Control}). 

To make field gradients explicit in the equation of motion, we use a position-dependent field model. The gradient captures how the field varies under a small displacement $\Delta x$ from an operating point. Using a first-order Taylor expansion, this yields (see also \cite{valdastri2023collaboRobotsPermMag}, \cite{petruska2020magMethodsRobot}): 
\begin{align}
     \b = |\b| \< \sin u_\alpha \\ \cos u_\alpha \> + \< u_{\mathrm{xx}} & u_{\mathrm{xy}} \\ u_{\mathrm{yx}} & u_{\mathrm{yy}} \> \< \Delta x \\ \Delta y \>, \label{eq:field_model_with_gradients}
\end{align}
where $u_{\mathrm{ab}} \coloneqq \frac{\partial b_a}{\partial b}$. For the case of the actuator, the left term represents the magnetic field vector at the center of the magnetic workspace and $\Delta \bm{x} = \ell_\mathrm{m} \< \sin(\alpha) & \cos(\alpha) \>^\T$. This results in a potential energy of the magnetic field, that depends on the magnetic field vector, the position (or more specific the orientation), and the gradients (additional control inputs). Rederiving the Lagrangian and the linearized dynamics  results in  
\begin{align*}
    J \ddot{\alpha} + d\dot{\alpha} + (-\eta g + \mb )  \alpha=  \mb u_\alpha \\ \ldots + \ell_m|\tilde{\boldsymbol{m}}| u_{x y}+\ell_m|\tilde{\boldsymbol{m}}| u_{y x}
\end{align*} 
Note that this approximation fails when gradients are large, as it can violate the field-alignment assumption. This highlights the challenge of embedding fields and gradients directly in the equations of motion, since doing so typically requires additional modeling assumptions. By contrast, forces are more straightforward to include because they relate directly to the resulting motion.

\section{Singular Value Decomposition}
\label{app:svd_torque_map}
The singular value decomposition (SVD) of the matrix $\M_{\b} = \skews{\m}$ provides valuable insights into the structure of magnetic torque generation. Recall that the physical relation $\torque = \m \times \b$ holds independently of the reference frame. In particular, in the body-fixed frame $\mathcal{B}$, we write $\torqueB = \mB \times \bB$. 

Since $\skews{\mB}$ is sparse, the SVD is more easily computed in the body-fixed frame and then mapped to the inertial frame using a rotation matrix. Let the frame transformation from the body-fixed frame to the inertial frame be given by the rotation matrix $\R \in \mathrm{SO}(3)$ such that any vector $\bm{z}$ transforms as $\bm{z} = \R^\top\Bprefix{\bm{z}}$.

Because singular values are invariant under rotation, the singular values of $\skews{\m}$ and $\skews{\mB}$ are identical and satisfy
$
    \bm{\Sigma} = \mathrm{diag}\{ \abs{\m}, \abs{\m}, 0 \}. 
$
The SVD of $\skews{\mB}$ can be obtained by inspection:
\begin{align}
    \skews{\mB} =\abs{\m} \< 0 & -1 & 0 \\ 1 & 0 & 0 \\ 0 & 0 & 0 \> = \Bprefix{\U} \bm{\Sigma} \Bprefix{\V}^\T, \\
    \Bprefix{\U} = \< 0 & -1 & 0 \\ 1 & 0 & 0 \\ 0 & 0 & 1 \>, \qquad
    \Bprefix{\V} = \I^{3 \times 3}. 
\end{align}
Transforming back to the inertial frame, we obtain the SVD of $\skews{\m}$ as:
\begin{align}
    \M &= \skews{\m} = \U \Sigma \V^\T, \\
    \U &= \R^\T \Bprefix{\U}, \qquad \V^\T = \R. 
\end{align}
This decomposition reveals several key properties:
\begin{itemize}
    \item The nullspace of $\skews{\m}$ is one-dimensional:
    \begin{align*}
        \mathrm{null}(\skews{\m}) = \spans{\m}
    \end{align*}
    implying that any magnetic field component parallel to the dipole $\m$ cannot contribute to torque generation.
    \item The image of $\skews{\m}$ is the plane spanned by the first two left singular vectors:
    \begin{align}
        \mathrm{image}(\skews{\m}) = \mathrm{span}\{ \bm{u}_1, \bm{u}_2 \}.
    \end{align}
    Note that both vectors satisfy $\bm{u}_1^\T \m = \bm{u}_2^\T \m = 0$, meaning they are both orthogonal to the dipole moment $\m$. Hence, only field components perpendicular to the dipole contribute to torque generation.
\end{itemize}

\section{Feasibility of torque-to-current mapping}
Our goal is to guarantee that, for a given torque $\torque$, there exists a current $\current \in \mathbb{R}^n$ such that the following equation can be satisfied exactly, i.e.,
$
\norm{\M_{\b} \Act_\b \current - \torque} = 0,
$
rather than merely in a least-squares sense. In other words, for a given $\torque \in \mathbb{R}^3$, we seek $\current \in \mathbb{R}^n$ such that
$
\M_\b \Act_\b \current = \torque.
$
A necessary and sufficient condition for such a solution to exist is
$
\torque \in \range{\M_\b \Act_\b}.
$
Since $\range{\M_{\b} \Act_\b} \subseteq \range{\M_{\b}}$, the condition $\torque \in \range{\M_\b}$ is necessary. Assuming the absence of magnetic field singularities, i.e., $\Act_\b$ has full row rank with $\mathrm{rank}(\Act_\b) = 3$, we have
$
\range{\M_\b \Act_\b} = \range{\M_\b},
$
and the condition is therefore also sufficient.

From the singular-value decomposition of $\M_\b = \skews{\m}$ (App.~\ref{app:svd_torque_map}), we know that
$
\range{\M_\b} = \spans{\m}^{\perp}
$
is the two-dimensional plane orthogonal to $\m$. Thus, by construction of the torque vector in the body-fixed frame as
$
\torqueB = \< \tau_{\c, x}, \tau_{\c, y}, 0 \>^\T,
$
we ensure $\torque \in \range{\M_\b}$, and therefore $\torque \in \range{\M_\b \Act_\b}$. Consequently, there always exists a $\current \in \mathbb{R}^n$ satisfying $\M_\b \Act_\b \current = \torque$ exactly.

Note that this feasibility guarantee applies specifically to the simplified torque-to-current mapping presented here. When including additional components such as gradient effects or the Jacobian transformation, the feasibility analysis becomes more complex. Nevertheless, this analysis provides important mathematical motivation for defining the torque in the body-fixed frame.

\section{Analysis of Torque Allocation Methods}
\label{app:allocation_comparison}
This section provides a detailed quantitative analysis of how the one-step and two-step allocation methods differ due to the presence of a nontrivial nullspace in the torque/force-to-current mapping. While we focus the discussion on the pure torque-to-current relation $
    \torque = \M_{\b} \Act_\b \current,
$
the same reasoning directly extends to the combined torque-force mapping, $\< \torque^\T & \force^\T \>^\T = \M \Act \current$ as well as to the formulation with the Jacobian $\jacobi$, i.e., $\torque_\c = \jacobi \M \Act \current$.
\begin{figure}
	\centering
	\includegraphics[scale=0.3, trim = 0.0cm 0.0cm 0.0cm 0.0cm, clip]{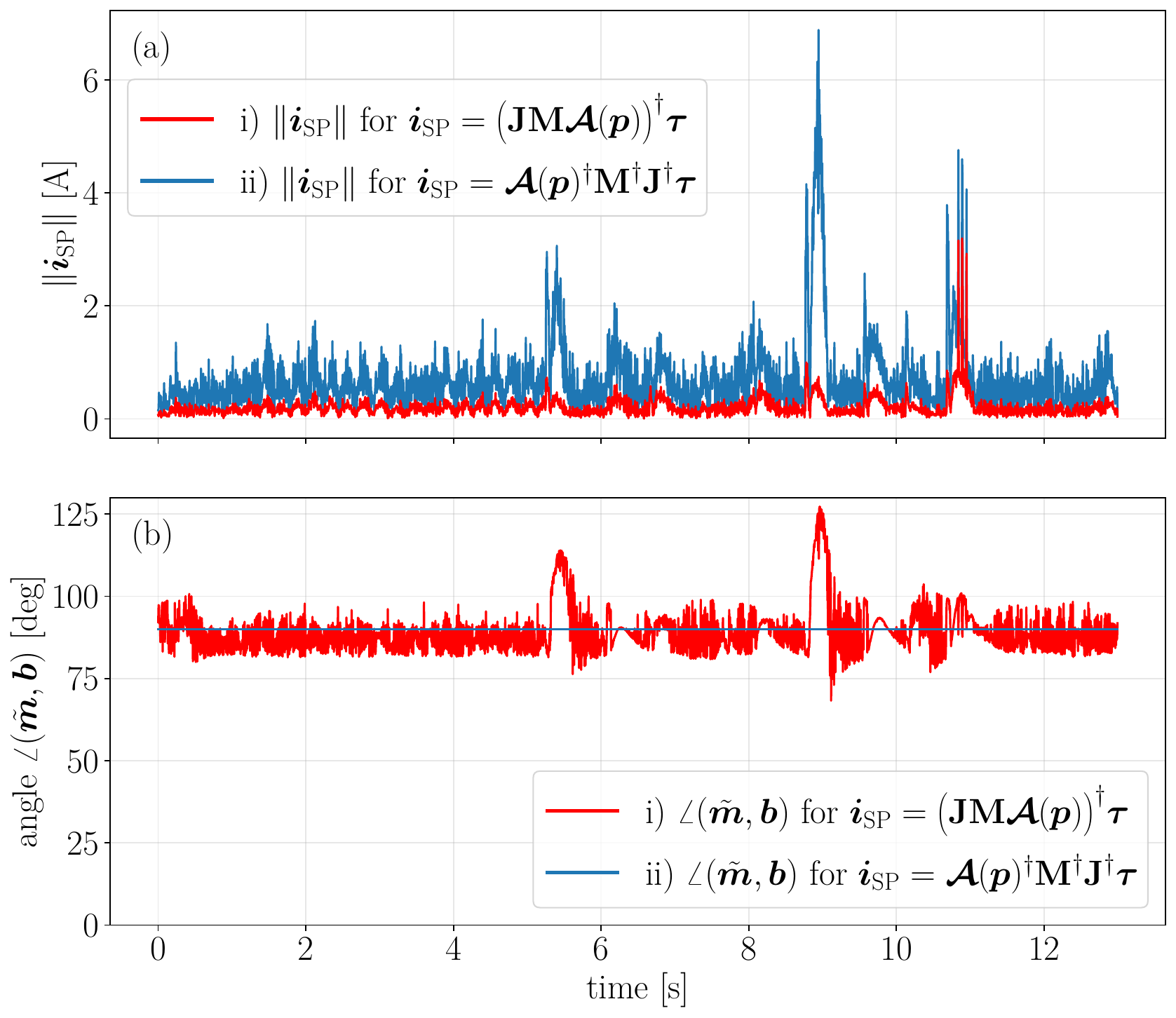}
	\caption{Comparison of current norms and field-dipole angles for different torque-to-current allocation strategies. The upper plot shows the $\ell^2$-norm of the coil currents required by the one-step allocation (a) and one representative multi-step allocation (b). The multi-step approach shown maps torques to forces/torques, then to fields/gradients, and finally to currents. Similar results are observed for the two alternative multi-step allocations -  $\current_\SP = \Act(\p)^\dagger \bigl(\jacobi \M\bigr)^\dagger \torque$ and $\current_\SP = \bigl(\M \Act(\p)\bigr)^\dagger \jacobi^\dagger \torque$ - but are omitted for clarity. The one-step method consistently yields the minimum-norm (i.e., minimum-energy) solution. The lower plot depicts the angle between the magnetic dipole moment $\m$ and the magnetic field $\b$. Multi-step allocation (b) enforces strict orthogonality between field and dipole, while the one-step method (a) allows small parallel components to reduce current demand. Data was recorded during inverted pendulum stabilization using the OctoMag eMNS with control based on (a); all other results were computed offline for comparison. Current spikes correspond to external disturbances where higher currents were needed for recovery.}
	\label{fig:torque_allocation_norm_comparison_app}
\vspace{\figcapskip}
\end{figure}

By definition of the Moore–Penrose pseudoinverse~\cite[Thm.~5.5.1]{golub2013matrix}, $\bm{x} = \bm{C}^\dagger \bm{d}$ is the minimum-norm least-squares solution to the linear system of equations $\bm{C} \bm{x} = \bm{d}$; that is, it minimizes the residual $\|\bm{C} \bm{x} - \bm{d}\|$, and among all minimizers, the solution has the smallest 2-norm. 
Hence, we can conclude:
\begin{itemize}
    \item In the \emph{one-step allocation}~\eqref{eq:onestep_alloc}, the optimization problem  
    \[
    \current^{\aI} \coloneqq \arg \min_{\current} \|\current\| \quad \text{s.t.} \quad \M_\b \Act_\b \current = \torque
    \]
    yields the \emph{minimum-energy} (minimum 2-norm current) solution for the given torque task.
    \item In the \emph{two-step allocation}~\eqref{eq:two_step_alloc}, the first stage solves  
    \[
    \b^{\aII} \coloneqq \arg \min_\b \|\b\| \quad \text{s.t.} \quad \M_\b \b = \torque,
    \]
    yielding the \emph{minimum field magnitude} solution. The second stage then maps that minimum-norm field to currents.  
\end{itemize}

To understand the relationship between the two allocations, we note that any solution to $\M_\b \bm{z} = \torque$ can be written as
$
    \bm{z} = \b + \zeta \m, \quad \zeta \in \mathbb{R},
$
where $\b$ is any valid particular solution, which we construct as $\b = \M_\b^\dagger \torque$ and $\zeta \m \in \nulls{\M_\b}$ is a parallel-field component that does not contribute to torque generation.  
For the two-step allocation, $\zeta = 0$ by construction.  

The one-step allocation solves  
\[
\min_{\zeta \in \mathbb{R}} \norm{\current(\zeta)}^2 =\min_{\zeta \in \mathbb{R}} \big\| \Act_\b^\dagger (\b + \zeta \m) \big\|,
\]
which is an unconstrained optimization problem, with the minimizer 
\begin{align}
    \zeta^* = -\frac{(\Act_\b^\dagger \b)^\T (\Act_\b^\dagger \m)}{\|\Act_\b^\dagger \m\|^2}.
\end{align}
This yields the minimum-norm current allocation
\begin{align}
    \current^\aI
    &= \current^\aII + \zeta^* \Act_\b^\dagger \m. \label{eq:one_step_vs_two_step_app}
\end{align}
The corresponding norms are related by
\begin{align}
    \|\current^\aI\|^2 &= \|\current^\aII\|^2 - \frac{\bigl((\Act_\b^\dagger \b)^\T (\Act_\b^\dagger \m)\bigr)^2}{\|\Act_\b^\dagger \m\|^2} 
\end{align}
which implies $\norm{\current^\aI} \leq \|\current^\aII\|$ with equality if and only if the numerator of the second term is zero.

Multiplying~\eqref{eq:one_step_vs_two_step_app} on the left by $\Act_\b$ and assuming $\Act_\b$ has full row rank, i.e., $\mathrm{rank}(\Act_\b) = 3$ (i.e. no magnetic singularity), gives the equivalent field relation:
\begin{align}
    \b^\aI = \b + \zeta^* \m.
    \label{eq:field_relation_app}
\end{align}
Note that $\b = \b^\aII$ is strictly orthogonal to $\m$, since $\b^\aII = \M_\b^\dagger \torque = \skews{\m}^\dagger \torque = -\abs{\m}^{-2} \skews{\m} \torque$ which results in $ \m^\T \b^\aII = 0$. Consequently, 
\begin{align}
    \norm{\b^\aI}^2 &= \norm{\b^\aII}^2 + \zeta^{*2} \norm{\m}^2 + 2 \zeta^*  \m^\T \b^{\aII} \\
    &= \norm{\b^\aII}^2 + \zeta^{*2} \norm{\m}^2,
\end{align}
which implies that $\|\b^\aI\| \geq \|\b^\aII\|$. Hence, we get the counterintuitive result that despite having a larger field norm $\|\b^\aI\| \geq \|\b^\aII\|$, the one-step allocation still requires smaller currents, $\|\current^\aI\| \leq \|\current^\aII\|$. 

The one-step allocation can exploit the nullspace of $\M_\b$ (parallel-field components) to lower the required coil currents (see \eqref{eq:field_relation_app}), while the two-step allocation strictly avoids any parallel component to minimize field strength.  
In practice, the parallel component in~\eqref{eq:field_relation_app} is typically small, so both methods yield fields that are nearly orthogonal to the dipole. That said, the two-step approach may be attractive in applications where minimizing field magnitude is the primary concern.

\section{Inter-Agent Magnetic Disturbance Estimation}
\label{app:magnetic_disturbance}

In multi-agent configurations, adjacent permanent magnetic agents exert parasitic coupling forces and torques on one another. Let $\mathbf{m}_i$ and $\mathbf{m}_j$ be the dipole moments of agents $i$ and $j$ at inertial positions $\mathbf{p}_i$ and $\mathbf{p}_j$. Defining the distance vector $\mathbf{r}_{ij} = \mathbf{p}_j - \mathbf{p}_i$ and its normalized direction $\hat{\mathbf{r}}_{ij} = \mathbf{r}_{ij} / \|\mathbf{r}_{ij}\|$, the magnetic torque $\boldsymbol{\tau}_{i \to j}$ and magnetic force $\mathbf{f}_{i \to j}$ exerted by agent $i$ on agent $j$ are given by the formulation in \cite{petruska2020magMethodsRobot}:
\begin{equation}
    \boldsymbol{\tau}_{i \to j} = \frac{\mu_0}{4\pi\|\mathbf{r}_{ij}\|^3} \skews{\mathbf{m}_j} \left(3\hat{\mathbf{r}}_{ij}\hat{\mathbf{r}}_{ij}^\top - \I\right) \mathbf{m}_i
    \label{eq:dist_torque}
\end{equation}
\begin{equation}
    \begin{split}
        \mathbf{f}_{i \to j} &= \frac{3\mu_0}{4\pi\|\mathbf{r}_{ij}\|^4} \Big( (\hat{\mathbf{r}}_{ij}^\top \mathbf{m}_j)\mathbf{m}_i + (\hat{\mathbf{r}}_{ij}^\top \mathbf{m}_i)\mathbf{m}_j + \ldots\\
        &\quad \ldots \left(\mathbf{m}_i^\top \mathbf{m}_j - 5(\hat{\mathbf{r}}_{ij}^\top \mathbf{m}_i)(\hat{\mathbf{r}}_{ij}^\top \mathbf{m}_j)\right)\hat{\mathbf{r}}_{ij} \Big).
    \end{split}
    \label{eq:dist_force}
\end{equation}
The actuator is constrained by a spherical joint and therefore the translational force in \eqref{eq:dist_force} induces a mechanical torque. Hence, the total disturbance torque is $\boldsymbol{\tau}_{\mathrm{d},i \to j} = \boldsymbol{\tau}_{i \to j} + (\ell_\mathrm{m} \mathbf{n}_j) \times \mathbf{f}_{i \to j}$, which we subsequently transform into the body-fixed frame. 

\section{Practical Considerations for $2\times3$D Pendulum}
\label{app:practical_considerations}

\paragraph{Allocation}
Although we include forces through the Jacobian in our single-task formulation, it is also possible to simplify and directly map torques to currents through $\current = \< \, \M_\b \, \Act(\p) \, \>^\dagger \, \torque_{\mathrm{c}}$ in the single-agent case. However, this force-ignoring approximation breaks down in multi-agent scenarios due to strong field gradients that induce destabilizing forces at the actuators. Incorporating the force component through the Jacobian $\jacobi(\alpha_i,\beta_i)$ thus becomes critical for achieving stable simultaneous control of multiple agents.

\paragraph{Offset Compensation}
As detailed in \Cref{sec:Experimental Platform}, our experimental implementation employs a decoupled LQRI control architecture with an independent controller for each degree of freedom. Integral action is crucial for eliminating the large steady-state errors that would otherwise persist. The stabilization procedure involves releasing each pendulum sequentially: one is stabilized while its integral term converges, and the process is then repeated for the second. Once calibrated, these integral values can be stored and reused as a warm-start, enabling the simultaneous release and stabilization of both pendulums in subsequent experiments.

Two primary factors contribute to the observed steady-state deflections. The first is the system's high sensitivity to misalignments between the gravitational vector and the motion-capture frame, which causes significant error amplification. This effect is analyzed and compensated for using the method in \cite{zughaibi2024balancing}, allowing us to isolate its contribution (see \Cref{fig:practical_considerations}). Consequently, the remaining deflections are likely attributed to inaccuracies in the underlying multipole expansion model used to derive $\Act(\p)$ and interactions between the permanent magnets. Further inaccuracies arise from the point-dipole assumption, as spatial field variations across the magnet's finite volume generate unmodeled forces and torques. 

\begin{figure*}
	\centering
	\includegraphics[scale=0.305,   trim = 0.0cm 0.0cm 0.0cm 0.0cm, clip ]{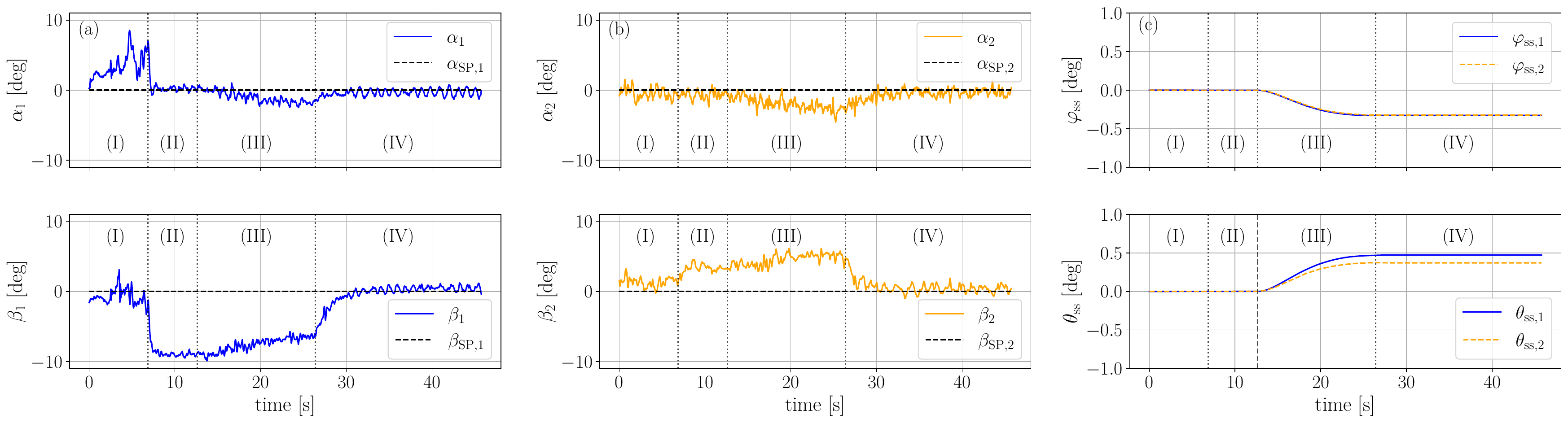} 
	\caption{The figure illustrates the behavior of the 2$\times$3D inverted pendulum system under different offset compensation strategies. Panels a) and b) show the trajectories of pendulums 1 and 2, respectively, while panel c) shows the learned deviation between the measured angles in the motion capture frame and the true gravitational direction. This deviation is used to correct the angle measurements with respect to gravity. Even small errors can lead to significant steady-state offsets, as analyzed in \cite{zughaibi2024balancing}, though this effect is more pronounced in single-agent scenarios - here, other sources of error dominate. We use this method to isolate sources of errors such that the resulting error after angle correction is likely attributed to inaccuracies in the magnetic model and permanent magnet interactions. Each pendulum is manually placed upright and released simultaneously (Phase I, at the first dashed line). In Phase II, no offset compensation is applied, leading to visible steady-state errors. In Phase III, only calibration-induced deviations from the motion capture system are compensated. In Phase IV, integral action is enabled to eliminate the remaining steady-state error. Note that it is not always possible to release both pendulums simultaneously due to large steady-state offsets (dependent on motion capture calibration), as collisions may occur. However, reliable stabilization can be achieved by sequentially releasing and stabilizing each pendulum with integral action, then storing the integral states for warm-starting future experiments.}
	\label{fig:practical_considerations}
\vspace{\figcapskip}
\end{figure*}

} 

\section*{Acknowledgments}
The authors would like to thank the team at MagnebotiX AG for their technical support and Neelaksh Singh for the insightful discussions. Additionally, the authors utilized ChatGPT, Gemini, and Claude for sentence restructuring to improve the readability of the manuscript.

Bradley Nelson is the co-founder of MagnebotiX AG and Nanoflex Robotics AG, which commercialize the OctoMag and Navion systems. The other authors declare no conflict of interest.

\vspace{-3mm}
\bibliographystyle{IEEEtran}
\bibliography{bibliography}

\begin{IEEEbiography}[{\includegraphics[width=1in,height=1.25in,clip,keepaspectratio]{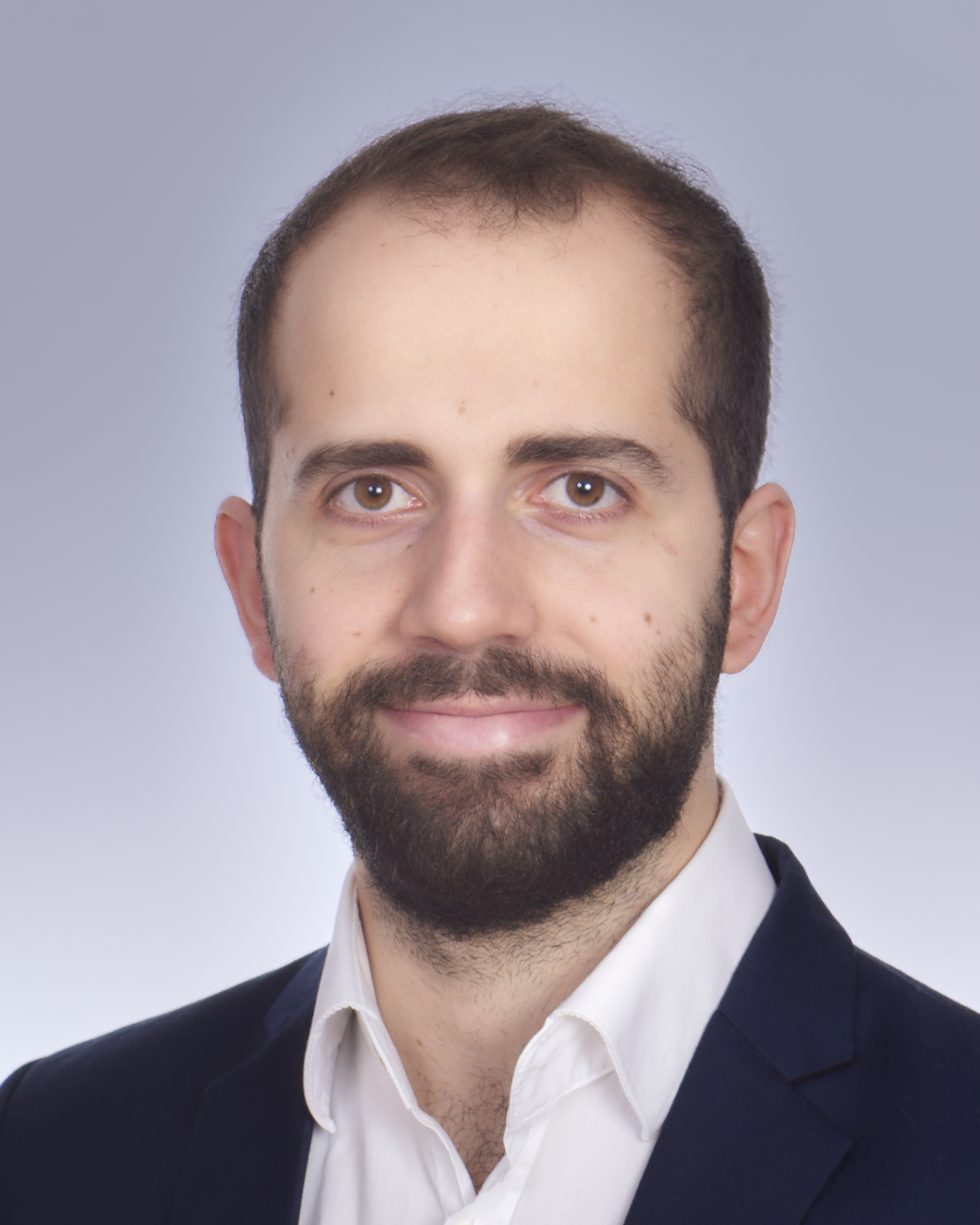}}]{Jasan Zughaibi} received the B.Sc. degree in Engineering Physics from TU Berlin, Germany, in 2017 and the M.Sc. degree in Robotics, Systems, and Control from ETH Zurich, Switzerland, in 2020, both with distinction. He is a doctoral student at the Multi-Scale Robotics Lab at ETH Zurich and a fellow of the Max Planck ETH Center for Learning Systems. For his master's thesis, he received the ETEL Master Award from ETH Zurich and the Hans-Eggenberger Award. His research interests include magnetic localization and actuation, dynamic feedback control, and online learning for control.
\end{IEEEbiography}

\begin{IEEEbiography}[{\includegraphics[width=1in,height=1.25in,clip,keepaspectratio]{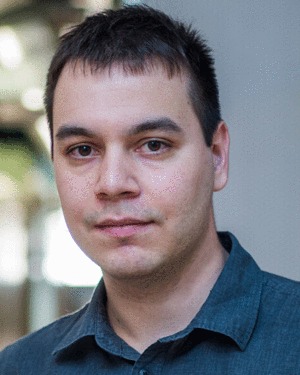}}]{Denis von Arx} received the B.Sc. and M.Sc. degrees in Electrical Engineering from ETH Zurich, Switzerland, in 2016 and 2018, respectively, and the Ph.D. degree in Robotics from the Multi-Scale Robotics Lab, ETH Zurich, in 2025. His research interests include electromagnetic actuation and localization for medical applications, power electronics for magnetic manipulation systems, and electric drives.
\end{IEEEbiography}

\begin{IEEEbiography}[{\includegraphics[width=1in,height=1.25in,clip,keepaspectratio]{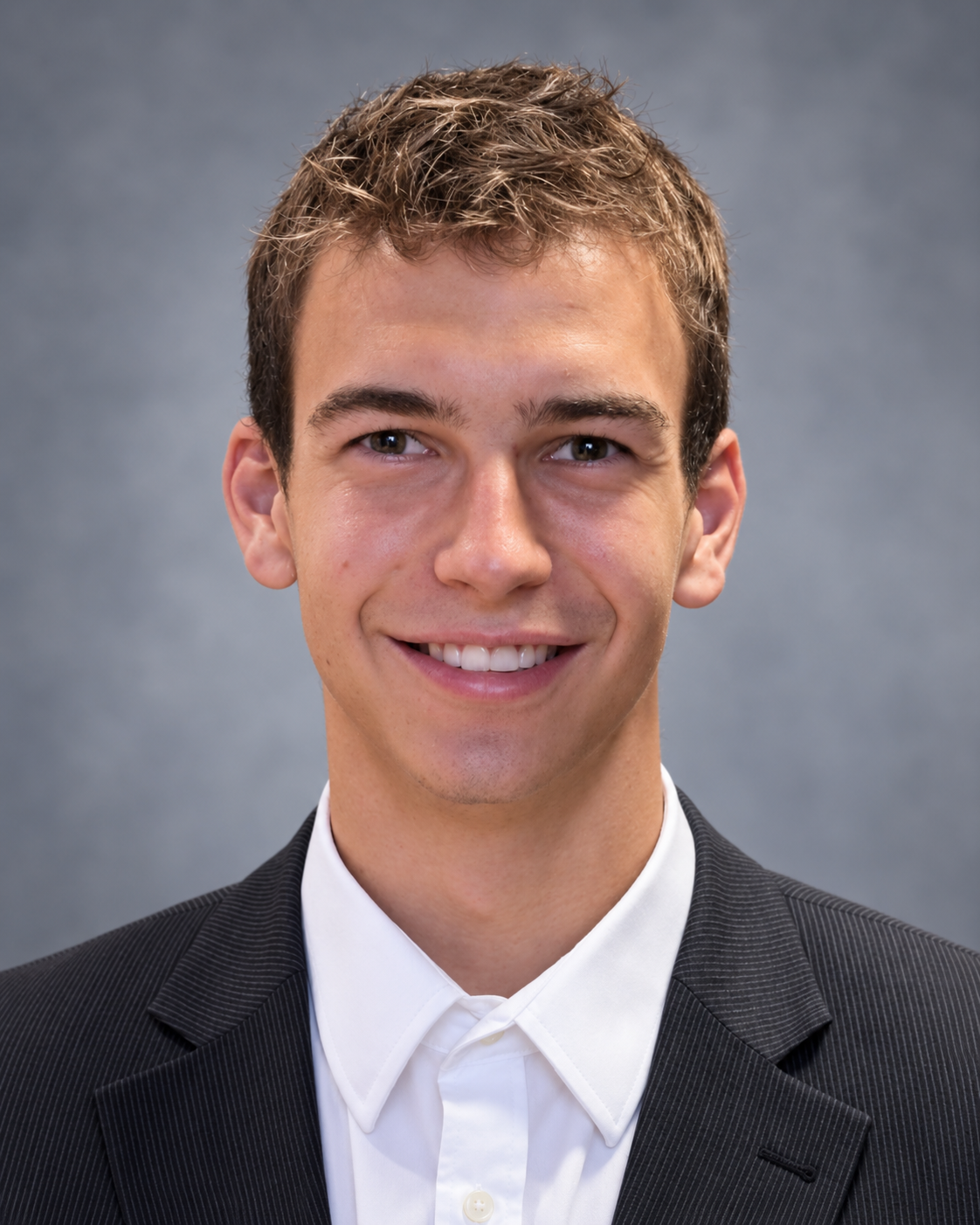}}]{Maurus Derungs} received the M.Sc. degree in Robotics, Systems, and Control from ETH Zurich, Switzerland. His research interests include learning, control, perception, and navigation for robotic and autonomous systems.
\end{IEEEbiography}

\begin{IEEEbiography}[{\includegraphics[width=1in,height=1.25in,clip,keepaspectratio]{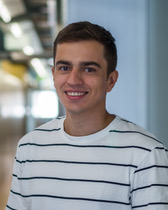}}]{Florian Heemeyer} received his B.Sc. and M.Sc. in Electrical Engineering and Information Technology from the Technical University of Munich (TUM) and his Ph.D. from ETH Zurich, where he conducted research in the Multi-Scale Robotics Lab. His research focused on robotic systems for minimally invasive surgery and telesurgery.
\end{IEEEbiography}

\begin{IEEEbiography}[{\includegraphics[width=1in,height=1.25in,clip,keepaspectratio]{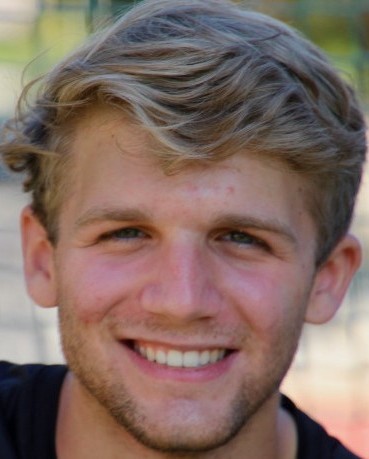}}]{Luca Antonelli} received his B.Sc. and M.Sc. degrees in mechanical engineering from ETH Zürich. His research interests include hardware design and control, sensor and actuator integration, novel electromechanical components, and algorithms for closed-loop motion control.
\end{IEEEbiography}

\begin{IEEEbiography}[{\includegraphics[width=1in,height=1.25in,clip,keepaspectratio]{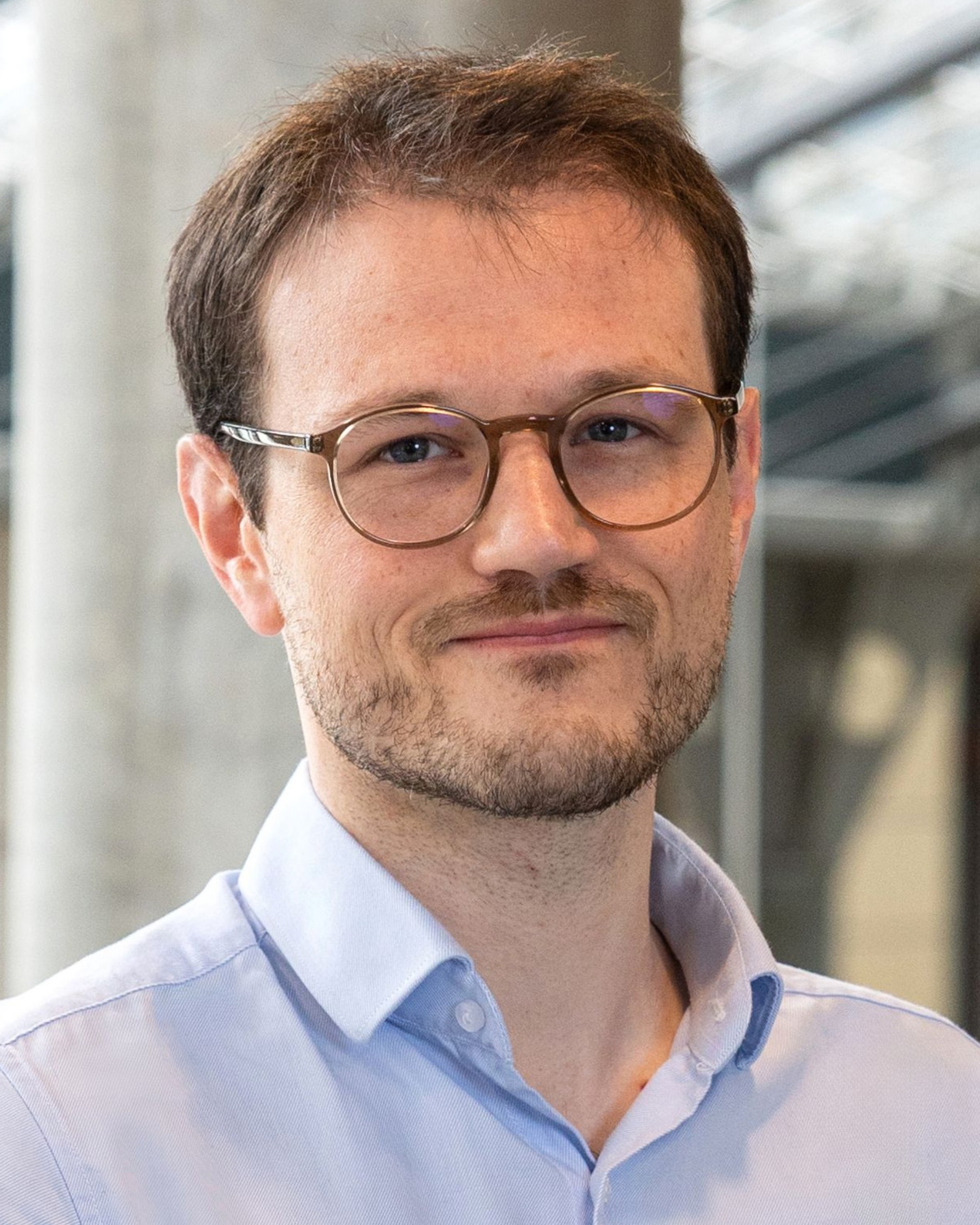}}]{Quentin Boehler} is an Assistant Professor of Robotics at ETH Zurich, where he leads the Medical Robotics Lab. He received an engineering degree in Mechatronics from INSA Strasbourg (2013), followed by an M.Sc. (2013) and a Ph.D. (2016) in Robotics from the University of Strasbourg. From 2017 to 2025, he was a researcher at ETH Zurich developing magnetic navigation technologies for medical applications. His research focuses on the design, control, and simulation of robotic systems for therapeutic and diagnostic interventions. He has co-authored over 50 publications and serves as an Associate Editor for IEEE Robotics and Automation Letters.
\end{IEEEbiography}

\begin{IEEEbiography}[{\includegraphics[width=1in,height=1.25in,clip,keepaspectratio]{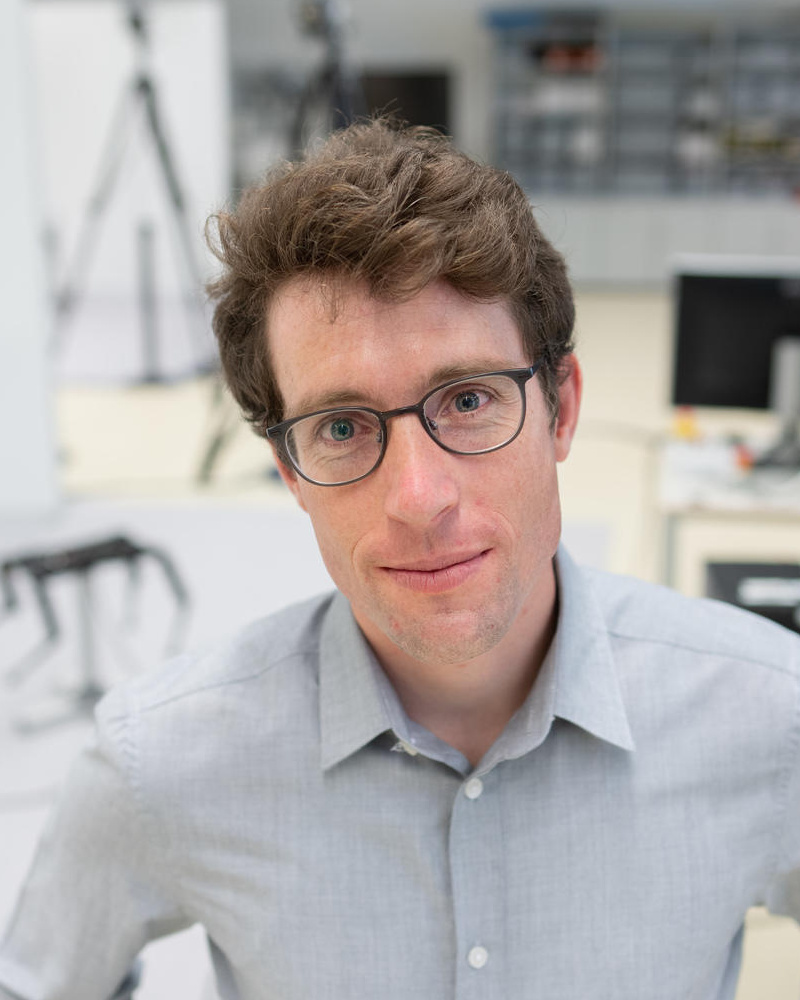}}]{Michael Muehlebach} obtained his Ph.D. at ETH Zurich under the supervision of Prof. R. D’Andrea in 2018. Since 2021, he has led an independent research group at the Max Planck Institute for Intelligent Systems, Tuebingen; since 2026, he has been a Full Professor at WU Vienna. His awards include the Outstanding D-MAVT Bachelor Award, Willi-Studer Prize for the best Master’s degree, ETH Medal and HILTI Prize for his Ph.D. research, Branco Weiss and Emmy Noether Fellowships, an Amazon research grant, and the IFAC Mechatronics Paper Prize. He also serves on the editorial board of Foundations and Trends in Machine Learning.
\end{IEEEbiography}

\begin{IEEEbiography}[{\includegraphics[width=1in,height=1.25in,clip,keepaspectratio]{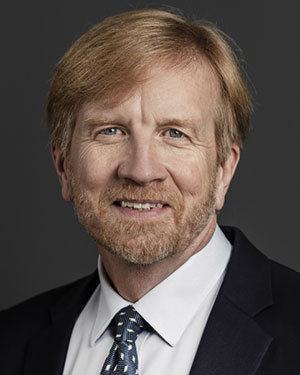}}]{Bradley J. Nelson} (Fellow, IEEE) received the B.S. degree in Mechanical Engineering from the University of Illinois, Urbana-Champaign, IL, USA, in 1984, the M.S. degree in Mechanical Engineering from the University of Minnesota, Minneapolis, MN, USA, in 1987, and the Ph.D. degree in Robotics from the School of Computer Science, Carnegie-Mellon University, Pittsburgh, PA, USA, in 1995. He has been Professor of Robotics and Intelligent Systems with ETH Zurich, Switzerland, since 2002. His research interests include micro and nanorobotics with applications in biology and medicine. He has received numerous awards for his work in robotics, nanotechnology, and biomedicine.
\end{IEEEbiography}

\vfill

\end{document}